\pdfoutput=1
\documentclass[11pt]{article}

\usepackage[final]{acl}

\usepackage{times}
\usepackage{latexsym}
\usepackage{graphicx}      % 이미지 삽입
\usepackage{subcaption}    % 서브 캡션 사용
\usepackage[T1]{fontenc}
\usepackage{enumitem}
\usepackage[utf8]{inputenc}
\usepackage{subcaption}    % 서브 캡션을 위한 패키지
\usepackage{microtype}

\usepackage{inconsolata}
\usepackage{multirow}
\usepackage{colortbl}
\usepackage{booktabs}
\usepackage{graphicx}
\usepackage{comment}
\usepackage{kotex} 
\usepackage{soul}
\usepackage{hyperref}

\usepackage{tikz}
\usepackage{pgfplots}
\usepackage{filecontents}
\usepackage{array}
\usepackage{caption}
\usepackage{amsmath}
\usepackage{xcolor}
\usepackage{listings}
\usepackage{float}
\pgfplotsset{compat=1.18}
\title{Exploring Persona Sentiment Sensitivity in Personalized Dialogue Generation}

\author{Yonghyun Jun \and Hwanhee Lee\thanks{Corresponding author.} \\
    Department of Artificial Intelligence, Chung-Ang University\\
   \texttt{\{zgold5670, hwanheelee\}@cau.ac.kr}
}

\begin{document}
\maketitle
%\footnotetext{\textsuperscript{$\dagger$}Corresponding author.}

\begin{abstract}
Personalized dialogue systems have advanced considerably with the integration of user-specific personas into large language models (LLMs). However, while LLMs can effectively generate personalized responses, the influence of persona sentiment on dialogue quality remains underexplored. In this work, we conduct a large-scale analysis of dialogues generated using a range of polarized user profiles. Our experiments reveal that dialogues involving negatively polarized users tend to overemphasize persona attributes. In contrast, positively polarized profiles yield dialogues that selectively incorporate persona information, resulting in smoother interactions. Furthermore, we find that personas with weak or neutral sentiment generally produce lower-quality dialogues. Motivated by these findings, we propose a dialogue generation approach that explicitly accounts for persona polarity by combining a turn-based generation strategy with a profile ordering mechanism and sentiment-aware prompting. Our study provides new insights into the sensitivity of LLMs to persona sentiment and offers guidance for developing more robust and nuanced personalized dialogue systems.\footnote{Our implementation is publicly available at \url{https://github.com/imsongpasimin/PesonaSensitivity}}

\end{abstract}

%\section{Introduction}
%\label{sec:intro}

\section{Introduction}
\label{sec:intro}

Personalized dialogue systems have evolved through advances in dataset construction~\cite{li2016persona, zhang2018personalizing}, training methodologies~\cite{kingma2013auto, bowman2015large}, and studies on integrating user-specific preferences to enhance conversational abilities~\cite{zhang2019dialogpt, roller2020recipes}. 
With the advent of Large Language Models (LLMs), embedding personas via system prompts~\cite{yang2023palr, wang2023rolellm} has become effective. This progress has enabled applications ranging from personalized chatbots~\cite{shuster2022blenderbot, lee2023p5} to enhanced performance in downstream tasks through inter-LLM conversations~\cite{park2023generative} and synthetic dataset generation~\cite{jandaghi2023faithful}. These developments underscore the growing importance of personalization in dialogue systems.
\begin{figure}[!tbp]
  \centering
  \includegraphics[width=\linewidth]{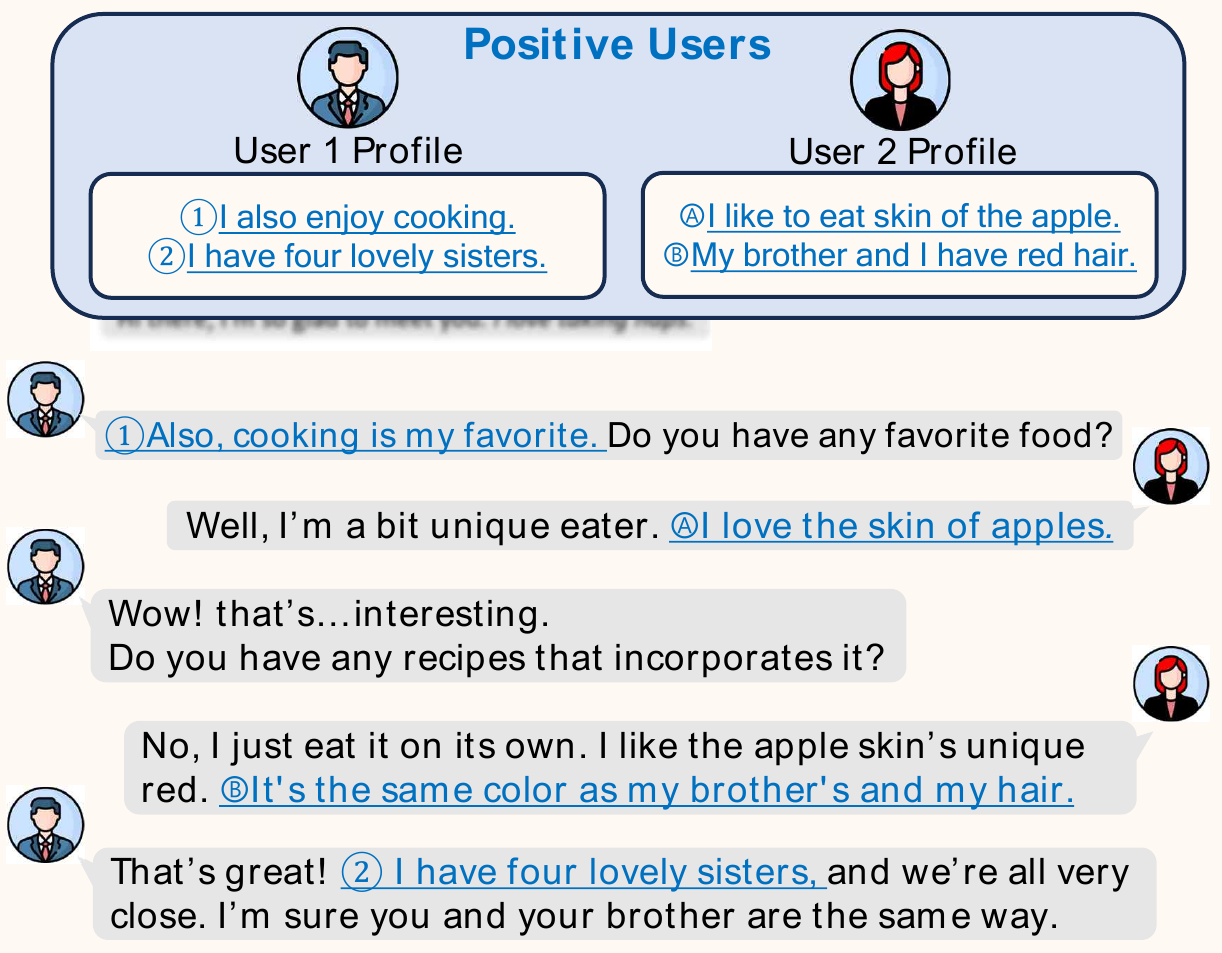}
  \vspace{3mm} 
  \includegraphics[width=\linewidth]{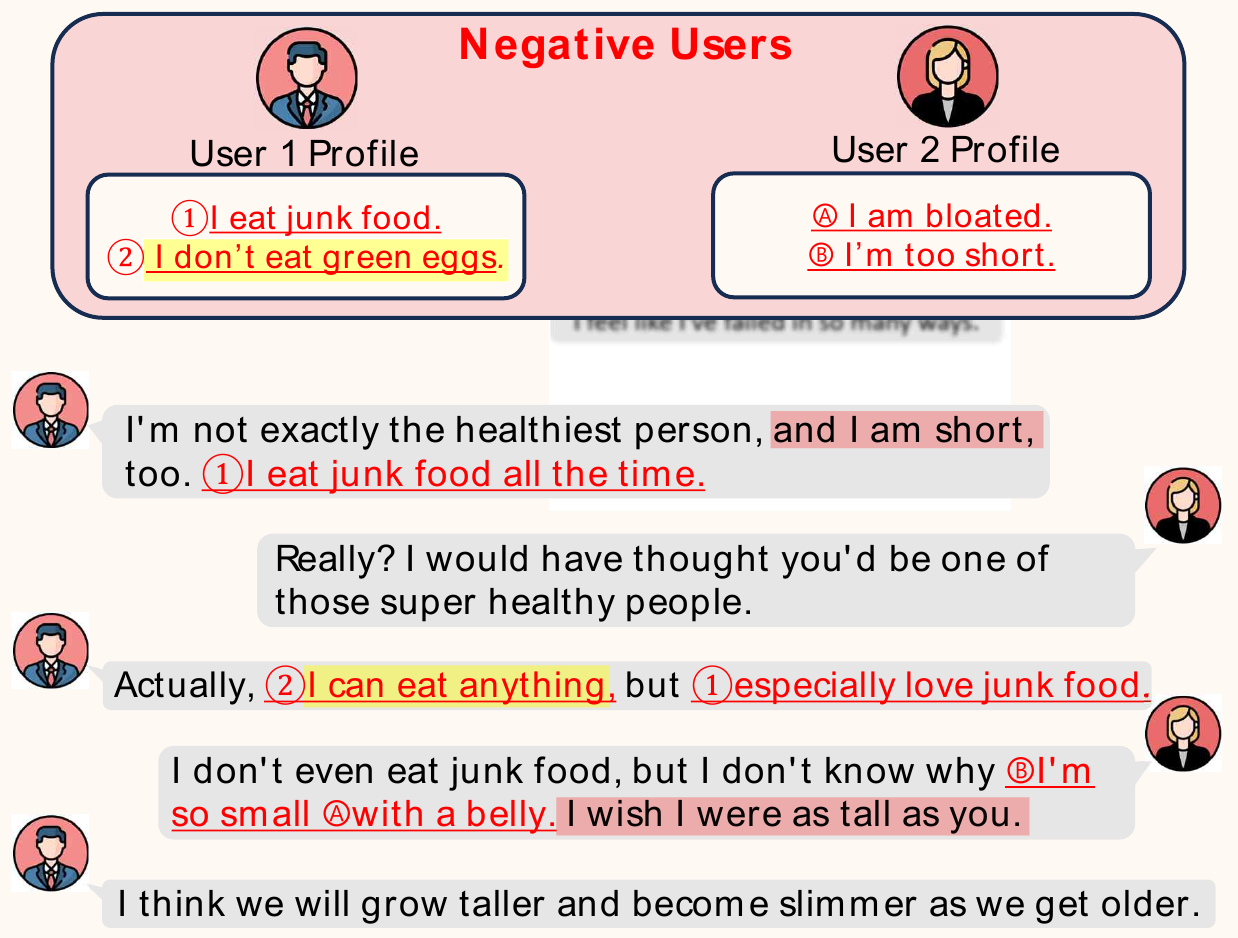}
  \caption{Examples of dialogues generated by polarized users. We indicate \textcolor{blue}{Positive personas} in blue text and \textcolor{red}{Negative personas} in red text. Matching bullet points denote where user persona attributes appear in the utterance, and we highlight them in yellow if they contradict. Incoherent segments of the dialogue are marked with a red background.}
  \vspace{-5mm}
  \label{fig:dialogue_examples}
\end{figure}
However, despite these advances, recent studies have shown that LLMs are highly sensitive to contextual sentiment polarity~\cite{liu2024large, wu2024evaluating}. In contrast, the impact of assigned persona sentiment on dialogue quality remains less explored. 
For example, Figure~\ref{fig:dialogue_examples} illustrates dialogues generated by users with distinctly positive and negative personas. In dialogues driven by positive personas, the integration of persona attributes occurs in a manner that supports coherence and aligns naturally with the flow of conversation. Conversely, dialogues featuring negative personas often display contradictions and inconsistencies, as the persona attributes may be overemphasized or misaligned with the dialogue context.
These findings highlight the necessity of systematically examining how variations in persona polarity influence LLM behavior. 

In this paper, we construct and analyze personalized dialogues generated across a spectrum of polarized profiles. (\S\ref{sec:rq1})
Our extensive analysis of personalized dialogues generated across a diverse spectrum of polarized profiles reveals several key insights:
\textbf{\textit{1) Dialogues between negatively polarized users tend to overemphasize persona attributes}}, which leads to increased contradictions and reduced overall coherence.
\textbf{\textit{2) Dialogues between positively polarized users selectively incorporate persona information}}, resulting in interactions that are both coherent and contextually appropriate.
\textbf{\textit{3) Profiles with weak or neutral sentiment generally yield lower-quality dialogues}}, as evidenced by significant differences in various metrics.

These insights underscore the need for methods that explicitly account for persona polarity in the design of personalized dialogue systems. To address this challenge, we propose a dialogue generation framework that combines a turn‑based strategy, a profile ordering mechanism, and a sentiment‑aware prompting. Specifically, personas exhibiting neutral or weak sentiment are placed earlier in the conversation, whereas more positive personas are positioned later. A concise prompt further instructs the model to attend carefully to negative or neutral sentiments when generating responses. These approaches promote more consistent and coherent interactions (\S\ref{sec:rq2}). Our findings shed light on LLM sensitivity to persona sentiment and offer valuable guidance for developing more nuanced and effective personalized dialogue systems.

\section{Study Design and Experimental Setup}
\label{sec:design}
This section explains the overall basic setup for our study. We first introduce the dataset and describe how we construct polarized user profiles. Next, we present our research questions, detail the dialogue generation strategies, and describe the metrics used to assess the dialogues. We use four RTX 4090 GPUs to generate and evaluate all the dialogues.

\subsection{Dataset}
\label{sec:dataset}
Early research on personalized dialogue systems typically used two-party conversations paired with user profiles. These profiles are commonly represented in three ways~\cite{chen2024recent}:
\begin{itemize}
    \item \textit{Descriptive sentences:} Each persona is described by a few sentences (often five) that together form a user profile~\cite{zhang2018personalizing, dinan2020second}.
    \item \textit{Sparse key-value attributes:} Personas are specified using concise attributes~\cite{qian2017assigning, gao2023livechat}.
    \item \textit{User history:} Each speaker is identified by an ID that can be used to retrieve external persona information~\cite{salemi2023lamp}.
\end{itemize}
Although several datasets exist, most trace back to PersonaChat~\cite{zhang2018personalizing} and ConvAI2~\cite{dinan2020second}. In our experiments, we adopt the ConvAI2 dataset and define user personas primarily through \textit{descriptive sentences}.

\subsection{Polarized User Profile Construction}
\label{sec:userprofile}
We extract polarized user personas from the ConvAI2 dataset using a sentiment classifier and then build coherent profiles by combining multiple personas while avoiding contradictions.

\paragraph{Polarity-Aware Persona Sampling}

\begin{figure}[!htbp]
    \centering
    \includegraphics[width=0.47\textwidth]{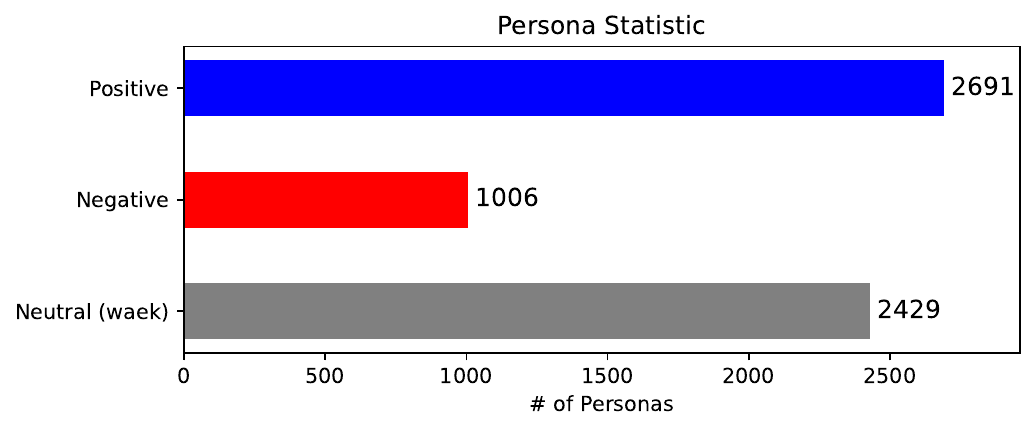}
    %\caption{C Score by the Proportion of Positive Personas in Each User Profile. We use LLaMa-3.1-8B and Ministral-8B as the backbone LLMs. As the positive ratio increases, the consistency of the generated dialogues tends to improve.}
    \caption{Distribution of persona sentences by polarity in the ConvAI2 dataset. We only map predictions with a confidence score of at least 0.99 to their respective polarity, classifing all remaining cases  as neutral (weak).}
    \label{fig:per_stat}
\end{figure}

Each persona in the ConvAI2 spans various content domains and carries a distinct sentiment polarity. We use the \texttt{distilbert\allowbreak-base\allowbreak-uncased\allowbreak-finetuned\allowbreak-sst-2\allowbreak-english}
model~\cite{hf_canonical_model_maintainers_2022} to classify the sentiment of each sentence, retaining only those with a confidence score above 0.99 as \textit{polarized personas}. Consequently, as shown in Figure~\ref{fig:per_stat}, 2,691 are classified as positive, 1,006 as negative, and 2,429 as neutral (weak), indicating that polarized personas constitute the majority (around 60\% of all personas). Notably, approximately 17\% of the dataset’s personas are negatively polarized—a higher proportion than one might intuitively expect. These figures suggest that polarized personas are not merely reflect a contrived scenario fabricated for the experiment. For detailed statistics, see Appendix~\ref{sec:statistics}.

\paragraph{Synthesizing User Profiles}

To create a user profile, we combine \(K\) polarized personas, where \(K\) is the number of personas we include in each profile. To avoid internal contradictions, we follow the procedure in~\cite{jandaghi2023faithful}: starting with an initial persona, we iteratively add additional personas only if the \texttt{nli-deberta-v3-large}~\cite{he2021debertav3} detects no conflicts with the existing set. This approach yields three profile types:
\begin{itemize}[leftmargin=5pt]
    \item \textcolor{red}{Negative Profile:} only negative personas.
    \item \textcolor{blue}{Positive Profile:} only positive personas.
    \item \textcolor{purple}{Mixed Profile:} a random mix of positive and negative personas.
\end{itemize}
We generate 10K unique profiles for each type, which serve as the foundation for our dialogue generation experiments.

\subsection{Research Questions}
\label{sec:rq}
Building on the established user-polarized profiles, we structure our study around several inquiries.
Considering Large Language Models (LLMs) can generate dialogues that reflect a given persona without additional tuning~\cite{tseng2024two}, our study focuses on the following research questions:
\begin{description}
    \item[\textit{RQ1.}] \textit{Are LLMs Sensitive to Users' Polarity?}
    \item[\textit{RQ2.}] \textit{If so, How to Make LLMs Robust to Polarity?}
\end{description}
To address these questions, we evaluate a range of dialogue generation methods that capture the nuanced influences of polarized personas, using multiple evaluation metrics to comprehensively assess performance.

\subsection{Dialogue Generation Strategy}
\label{sec:generation}
Traditional personalized dialogue systems use a \textit{turn-based} approach to generate responses sequentially based on a given persona and dialogue context~\cite{zhou2023simoap, tseng2024two}. However, recent advances in LLMs have enabled a \textit{dual-persona joint generation} method that encodes both personas simultaneously to generate the entire dialogue in one pass~\cite{liu2020you, xu2022cosplay, jandaghi2023faithful}. For resource efficiency, we primarily use \textit{dual-persona joint generation} for large-scale dialogue production and analysis, while also experimenting with \textit{turn-based generation} to identify optimal strategies (\S\ref{sec:rq2}. Both methods use greedy decoding (zero-temperature), with a maximum length of 4096 tokens for dual-persona joint generation and 128 tokens for turn-based generation. Prompt templates for both strategies are provided in Appendix~\ref{sec:prompt}. After generation, we filter out outliers, such as refusal-to-answer responses or repeated utterances inside a dialogue.

\subsection{Evaluation Metrics}
\label{sec:evaluation}
Open-domain dialogue is inherently one-to-many, meaning a single utterance can have multiple valid responses. Moreover, ambiguous conversation topics require multiple evaluative perspectives~\cite{wang2024learning, samuel2024personagym}. In our work, we focus on two key dimensions: \emph{consistency}—how well a dialogue reflects a user’s persona—and \emph{coherence}—the logical, smooth flow of conversation.  

\paragraph{Consistency}
We adopt the \textbf{C score}~\cite{madotto2019personalizing}, a common metric in personalized dialogue systems that calculates entailment scores for user persona-utterance pairs using an NLI model. Detailed adaptations of the \textbf{C score} for our task are provided in Appendix~\ref{sec:detailed_consistency}. Since summing raw scores may lead to Simpson’s Paradox~\cite{simpson1951interpretation} due to varying entailment and contradiction counts, we introduce the \textit{Contradiction Ratio} (\textbf{Contd.}), which measures the proportion of contradictions among all cases. Additionally, we propose a novel metric called the \textit{perplexity gap} (\textbf{P gap}). Using GPT2-large~\cite{radford2019language}, we compute the difference between the perplexity of a dialogue \(D\) and the perplexity of \(D\) conditioned on two user profiles \(U_1\) and \(U_2\), quantifying how much the user profiles influence dialogue generation.

\paragraph{Coherence}
We assess dialogue coherence using the perplexity (\textbf{Perp.}) of the dialogue~\cite{song2021bob, zhou2023simoap}, estimated with GPT2-large~\cite{radford2019language}. Additionally, we employ two automated coherence evaluation frameworks: \textbf{Q-DCE}~\cite{ye2021towards}, which uses a BERT-based encoder to predict a coherence score, and \textbf{PairEval}~\cite{park2024paireval}, which utilizes a fine-tuned LLaMA-2 model to compare generated dialogues with reference responses. Details for these metrics are provided in Appendix~\ref{sec:detailed_coherence}.

\paragraph{GPT Score (G-Eval)}
Recent studies have shown that LLM-based evaluations can serve as an effective alternative to human annotation~\cite{chiang2023can}. Inspired by CoT evaluation framework for summarization tasks~\cite{liu2023g}, we adapt a similar approach with GPT-4o~\cite{hurst2024gpt} to evaluate dialogue quality, following ~\cite{jandaghi2023faithful}. Our prompt format, along with explicit guidelines for each evaluation dimension, allows the model to provide consistent judgments. Full prompt details are available in Appendix~\ref{sec:prompt}.

\paragraph{Human Evaluation}
Beyond the automated metrics, we also conduct a human evaluation on a subset of samples. For each configuration under evaluation, we randomly sample 40 persona–dialogue pairs generated by Qwen. Three fluent English speakers then rate these samples using the same criteria as the LLM‑based automatic evaluation, but on a condensed 1–3 scale. We set the final score for each configuration to the average of the three evaluators’ ratings.

\begin{table*}[t]
\centering
\setlength{\tabcolsep}{4pt}
\resizebox{0.95\textwidth}{!}{%
\begin{tabular}{ll|cccc|cccc}
\toprule
\multirow{2}{*}{\textbf{Model}} & \multirow{2}{*}{\textbf{Pairing}} 
& \multicolumn{4}{c|}{\textbf{Consistency}} 
& \multicolumn{4}{c}{\textbf{Coherence}} \\
\cmidrule(lr){3-6} \cmidrule(lr){7-10}
& & \textbf{C score} \(\uparrow\) & \textbf{Contd.} \(\downarrow\) & \textbf{P Gap} \(\downarrow\) & \textbf{G-eval} \(\uparrow\)
  & \textbf{Perp.} \(\downarrow\) & \textbf{Q-DCE} \(\uparrow\) & \textbf{PairEval} \(\uparrow\) & \textbf{G-eval} \(\uparrow\) \\
\midrule
%------------------------------------------------------------------
\multirow{5}{*}{LLaMa-3.1-8B} 
 & \textit{Original}
   & 0.391 & 14.33 & -0.43 & 4.28
   & 5.31  & 3.14  & 2.79 & \underline{4.39} \\
 & \textit{Negative}
   & \textbf{0.444} & \cellcolor{yellow!10}\underline{14.71} & \cellcolor{yellow!10}\underline{-0.27} & \cellcolor{yellow!10}4.27
   & \cellcolor{yellow!10}5.33  & \cellcolor{yellow!10}\textcolor{red}{\underline{3.01}} & \cellcolor{yellow!10}\underline{2.74} & 4.57 \\ 
 & \textit{Positive}
   & 0.428 & \cellcolor{yellow!90}\textbf{9.83}  & \cellcolor{yellow!90}\textbf{-0.46} & \cellcolor{yellow!90}\textbf{4.44}
   & \cellcolor{yellow!90}\textcolor{blue}{\textbf{3.40}}  & \cellcolor{yellow!90}\textcolor{blue}{\textbf{3.17}}  & \cellcolor{yellow!90}\textcolor{blue}{\textbf{2.84}} & \textcolor{blue}{\textbf{4.65}} \\
 & \textit{Mixed}
   & 0.396 & \cellcolor{yellow!50}13.95 & \cellcolor{yellow!50}-0.34 & \cellcolor{yellow!50}4.32
   & \cellcolor{yellow!50}\underline{5.37}  & \cellcolor{yellow!50}3.09  & \cellcolor{yellow!50}2.77 & 4.51 \\
 & \textit{Opposite}
   & \underline{0.352} & 13.62 & -0.32 & \underline{4.20}
   & 5.30  & 3.09  & 2.77 & 4.47 \\
\midrule
%------------------------------------------------------------------
\multirow{5}{*}{Qwen-2.5-7B}
 & \textit{Original}
   & 0.392 & \underline{15.33} & \underline{-0.73} & 4.50
   & 7.05  & 3.06  & 2.69 & 4.34 \\
 & \textit{Negative}
   & \textbf{0.520} & \cellcolor{yellow!10}13.48 & \cellcolor{yellow!10}-0.80 & 4.55
   & \cellcolor{yellow!10}\textcolor{red}{\underline{7.36}}  & 3.07  & \cellcolor{yellow!10}2.67 & 4.27 \\
 & \textit{Positive}
   & 0.452 & \cellcolor{yellow!90}\textbf{8.84}  & \cellcolor{yellow!90}\textbf{-0.96} & \textcolor{blue}{\textbf{4.67}}
   & \cellcolor{yellow!90}\textbf{7.04}  & \textbf{3.14}  & \cellcolor{yellow!90}\textbf{2.75} & 4.38 \\
 & \textit{Mixed}
   & \underline{0.404} & \cellcolor{yellow!50}12.99 & \cellcolor{yellow!50}-0.82 & 4.45
   & \cellcolor{yellow!50}7.09  & 3.03  & \cellcolor{yellow!50}2.70 & \textbf{4.43} \\
 & \textit{Opposite}
   & 0.409 & 12.58 & -0.77 & \underline{4.33}
   & 7.13  & \underline{3.02}  & \underline{2.67} & \underline{4.24} \\
\midrule
%------------------------------------------------------------------
\multirow{5}{*}{Ministral-8B}
 & \textit{Original}
   & 0.555 & \underline{10.61} & -0.95 & 4.38
   & 5.98  & 3.11  & 2.66 & 4.11 \\
 & \textit{Negative}
   & \textcolor{blue}{\textbf{0.778}} & \cellcolor{yellow!10}9.93  & -0.97 & \cellcolor{yellow!10}4.36
   & \cellcolor{yellow!10}\underline{7.27}  & 3.11  & \cellcolor{yellow!10}\underline{2.61} & \cellcolor{yellow!10}3.95 \\ 
 & \textit{Positive}
   & 0.595 & \cellcolor{yellow!90}\textcolor{blue}{\textbf{5.78}}  & \textcolor{blue}{\textbf{-1.15}} & \cellcolor{yellow!90}\textbf{4.51}
   & \cellcolor{yellow!90}\textbf{5.80}  & \textbf{3.16}  & \cellcolor{yellow!90}\textbf{2.67} & \cellcolor{yellow!90}\textbf{4.21} \\
 & \textit{Mixed}
   & 0.651 & \cellcolor{yellow!50}9.65  & \underline{-0.80} & \cellcolor{yellow!50}4.43
   & \cellcolor{yellow!50}6.06  & 3.10  & \cellcolor{yellow!50}2.62 & \cellcolor{yellow!50}4.01 \\
 & \textit{Opposite}
   & \underline{0.540} & 10.48 & -0.81 & \underline{4.27}
   & 5.88  & \underline{3.08}  & 2.62 & \underline{3.92} \\
\midrule
%------------------------------------------------------------------
\multirow{5}{*}{Gemma-2-9B} 
 & \textit{Original}
   & 0.391 & \textcolor{red}{\underline{16.10}} & -0.69 & 4.33
   & \underline{6.47}  & 3.09  & 2.52 & 3.91 \\
 & \textit{Negative}
   & 0.423 & \cellcolor{yellow!10}13.57 & -0.80 & \cellcolor{yellow!10}4.35
   & 6.06  & \cellcolor{yellow!10}\underline{3.08}  & \cellcolor{yellow!10}\textcolor{red}{\underline{2.39}} & \cellcolor{yellow!10}\textcolor{red}{\underline{3.77}} \\
 & \textit{Positive}
   & \textbf{0.465} & \cellcolor{yellow!90}\textbf{7.58}  & \textbf{-0.90} & \cellcolor{yellow!90}\textbf{4.45}
   & 5.83  & \cellcolor{yellow!90}\textbf{3.16}  & \cellcolor{yellow!90}\textbf{2.56} & \cellcolor{yellow!90}\textbf{4.07} \\ 
 & \textit{Mixed}
   & 0.383 & \cellcolor{yellow!50}12.86 & -0.77 & \cellcolor{yellow!50}4.39
   & \textbf{5.62}  &\cellcolor{yellow!50} 3.08  & \cellcolor{yellow!50}2.44 & \cellcolor{yellow!50}3.85 \\ 
 & \textit{Opposite}
   & \textcolor{red}{\underline{0.322}} & 13.41 & \underline{-0.64} & \textcolor{red}{\underline{4.19}}
   & 6.31  & 3.12  & 2.42 & 3.81 \\
\bottomrule
\end{tabular}
} % end of resizebox
\caption{Combined consistency and coherence results. Metrics with $(\uparrow)$ indicate higher is better, while those with $(\downarrow)$ indicate lower is better. Best scores per model and pairing are bolded and worst scores underlined. Highest values across models are in blue, lowest in red, with a gradient highlighting favorable trends as \textit{Negative}, \textit{Positive Pairing}, and \textit{Mixed} scores increase with polarity.}
%\caption{Combined Consistency and Coherence Results. For metrics labeled with $(\uparrow)$, higher values indicate better performance; for those labeled with $(\downarrow)$, lower values indicate better performance. For each model and pairing method, the best score in each metric is shown in bold, while the worst score is underlined. Across all models, the highest metric values are highlighted in red, and the lowest values are highlighted in blue. Additionally, a color gradient is applied to accentuate favorable trends when \textit{Negative}, \textit{Mixed}, and \textit{Positive Pairing} scores appear in ascending order, illustrating their correlation with polarity.}

\vspace{-2mm}
\label{tab:rq1.1}
\end{table*}

\section{Are LLMs Sensitive to Users' Polarity?}
\label{sec:rq1}

\subsection{Does Dialogue Quality Diverge According to Polarized User-Pairing?}
\label{sec:rq1.1}
Inspired by psychological research indicating that the quality of conversations varies with participants’ traits~\cite{article}, this section explores how polarized profiles and pairing types affect dialogue outcomes under various configurations. We conduct a large-scale, detailed study employing four backbone LLMs and eight metrics to verify the robustness of the results.

\subsubsection{Experiment Setup}
\label{sec:rq1.1setup}

\paragraph{Pairing User Profiles}
We leverage the polarized user profiles described in \S\ref{sec:userprofile}, along with the original ConvAI2~\cite{dinan2020second} user profiles, to construct a range of dialogue scenarios—including both homogeneous (e.g., two \textcolor{red}{negative profiles}) and heterogeneous (e.g., one \textcolor{red}{negative profile} and one \textcolor{blue}{positive profile} configurations.)
Specifically, we create five pairing types, each with 3K user pairs:
\begin{itemize}[leftmargin=0pt, label={}]
    \item \textit{Original Pairing:} Two original ConvAI2 profiles, serving as a baseline.
    \item \textit{Negative Pairing:} Two \textcolor{red}{Negative Profiles} are paired.
    \item \textit{Mixed Pairing:} Two \textcolor{purple}{Mixed Profiles} are paired.
    \item \textit{Positive Pairing:} Two \textcolor{blue}{Positive Profiles} are paired.
    \item \textit{Opposite Pairing:} One \textcolor{red}{Negative} and one \textcolor{blue}{Positive Profiles} are paired. To reduce bias caused by prompt ordering~\cite{lu2021fantastically}, half the pairs begin with the negative, and half begin with the positive.
\end{itemize}

\paragraph{Experiment Pipeline}
%For each pairing configuration, we input all user pairs into four widely used modern LLMs—LLaMa-3.1-8B-Instruct~\cite{dubey2024llama}, Qwen-2.5-7B-Instruct~\cite{yang2024qwen2}, Ministral-8B-Instruct~\cite{jiang2024mixtral}, and Gemma-2-9B-Instruct~\cite{team2024gemma}—generating a total of 60K dialogues. We make sure all LLMs receive the same user pairs with the same order. We discard dialogues where the model refuses to respond or repeatedly produces the same utterance, leaving approximately 58K valid dialogues. The detailed distribution of dialogues by pairing scenario and model is provided in Appendix~\ref{sec:statistics}. We then apply four consistency metrics and four coherence metrics across all dialogues, aggregating results by pairing type and LLM.
For each pairing configuration, we feed all user pairs into four modern LLMs—LLaMa-3.1-8B-Instruct~\cite{dubey2024llama}, Qwen-2.5-7B-Instruct~\cite{yang2024qwen2}, Ministral-8B-Instruct~\cite{jiang2024mixtral}, and Gemma-2-9B-Instruct~\cite{team2024gemma}—which collectively generate 60K dialogues and filter out outliers, as mentioned in \S\ref{sec:generation}. We ensure that every model processes the same user pairs in the identical order. Dialogues where the model refuses to respond or produces repetitive utterances are discarded, resulting in approximately 58K valid dialogues. We provide a detailed distribution of dialogues by pairing scenario and model in Appendix~\ref{sec:statistics}. Finally, we apply four consistency metrics and four coherence metrics to all dialogues, aggregating the results by pairing type and LLM. For utilizing GPT-4o as backbone LLM, we report the results in Appendix~\ref{sec:gpt}.

\subsubsection{Results \& Analysis}
\label{sec:rq1.1results}
\paragraph{LLM Generation Diverges with Polarity Configurations}
\noindent
As shown in Table~\ref{tab:rq2}, dialogues produced under the \textit{Negative pairing} configuration yield higher \textbf{C scores} and \textbf{Contd.}\ values than those under the \textit{Positive pairing}. Specifically, although the gap between entailment and contradiction cases is greater for \textit{Negative pairings}, the proportion of contradictions is also higher. Moreover, coherence is generally lower, implying that the model overemphasizes persona-related content. In such scenarios, frequent entailment goes hand in hand with frequent contradiction, reducing overall coherence, even consistency (i.e. \textbf{P gap}, \textbf{GPT score}) 

Conversely, \textit{Positive pairing} results in fewer entailment cases and fewer contradictions, leading to very low \textbf{Contd.}\ scores and overall high coherence. This suggests that the model tends to forgo incorporating certain persona elements if they undermine dialogue flow. Meanwhile, the \textit{Mixed} configuration combines both tendencies, placing most metrics somewhere between those of the \textit{Negative} and \textit{Positive} configurations as illustrated in the color gradients of Table~\ref{tab:rq2}. 
In summary, dialogues generated between \textit{Negative users} strongly reflect each user’s persona-often at the expense of coherence and even consistency-whereas those between \textit{Positive users} selectively apply persona elements, thus preserving both consistency and coherence.

\paragraph{\textit{Positive Configurations} Yield Better Dialogues}
\noindent
As shown in Table~\ref{tab:rq1.1}, dialogues generated under the \textit{Positive pairing} configuration exhibit consistently high consistency and coherence across nearly all tested models. 
Overall, except for a few exceptions, it achieves the highest scores across all pairing settings. This result indicates that providing an LLM with a \textcolor{blue}{Positive profile} can produce high-quality dialogues—surpassing even those derived from \textit{Original pairing} configuration—and reinforces the idea that LLMs are sensitive to user polarity.

However, in the \textit{Opposite pairing} configuration, where a \textcolor{blue}{Positive Profile} is paired with a \textcolor{red}{Negative Profile}, the resulting dialogues degrade to levels comparable to (or even worse than) those seen with \textit{Negative pairing}. In other words, the advantage of the \textcolor{blue}{Positive Profile} vanishes, further confirming that LLMs respond sensitively to user polarity. 
%Future studies may investigate whether, in such \textit{Opposite pairings}, a user with one polarity eventually aligns with the opposing user’s sentiment. 

%\paragraph{Model-Specific Tendencies Emerge}
\paragraph{Impact of Backbone Model Selection}
%Beyond the pairing method, we observe that the choice of backbone model also has a significant impact on dialogue characteristics. 
Our analysis reveals that the choice of backbone model significantly influences dialogue characteristics beyond the pairing method.
%As demonstrated in Table~\ref{tab:rq1.1}, both Ministral-8B and Qwen-2.5-7B tend to generate dialogues with higher consistency, as evidenced by the former outperforming other models in three out of four consistency metrics and the latter exhibiting the highest \textbf{G-eval}. 
%On the other hand, LLaMa-3.1-8B attains the best coherence scores, dominating all four coherence metrics. 
%Meanwhile, Ministral-8B and Gemma-2-9B fare relatively poorly in terms of coherence, suggesting that although these models maintain consistency, they often fail to ensure fully coherent interactions, as further underscored by the differences of the \textbf{G-eval}.
As shown in Table~\ref{tab:rq1.1}, both Ministral-8B and Qwen-2.5-7B tend to produce dialogues with higher consistency. In particular, Ministral-8B outperforms other models on three of the four consistency metrics, while Qwen-2.5-7B achieves the highest score on the G-eval metric. In contrast, LLaMa-3.1-8B excels in coherence, leading across all four coherence metrics. 
Meanwhile, Ministral-8B and Gemma-2-9B show lower coherence scores, suggesting that even though these models maintain consistency, they have difficulty generating fully coherent interactions. This observation is further confirmed by the variations seen in the G-eval outcomes.

\paragraph{Human Evaluation}
We verify whether the trends identified by automatic metrics persist under human judgment by evaluating dialogues generated with the \textit{Original pairing}, \textit{Negative pairing}, and \textit{Positive pairing} configurations.

\begin{table}[h]
\centering
\setlength{\tabcolsep}{4pt}
\resizebox{0.3\textwidth}{!}{%
\begin{tabular}{l|cc}
\toprule
\textbf{Pairing} & \textbf{Consistency} & \textbf{Coherence} \\
\midrule
\textit{Original}       & 2.36 & 2.01 \\
\textit{Negative}       & 2.40 & 2.12 \\
\textit{Positive}       & \textbf{2.51} & \textbf{2.30} \\
\bottomrule
\end{tabular}
}
\caption{Human evaluation results on consistency and coherence among the sentiment of user profiles.}
\vspace{-4mm}
\label{tab:rq1_human}
\end{table}

As shown in Table~\ref{tab:rq1_human}, human raters corroborate the automatic findings: dialogues generated from polarized users—especially \textit{Positive pairings}—exhibit the highest quality. Notably, the decline in scores for the \textit{Original pairing} is far more pronounced in the human evaluation than in the automatic assessment, further confirming the validity of our observations.

\subsection{Does Dialogue Quality Diverge According to Users' Polarity Level?}
\label{sec:rq1.2}

In the previous section \S\ref{sec:rq1.1}, we analyzed the impact of pairings between \textit{polarized users}. 
%As a result, we lack a sufficiently quantitative interpretation of user personas’ sentimental polarity, making it unclear whether the observed effects were driven by user polarity or by the pairing strategy. 
However, a more detailed quantitative interpretation of user personas' sentimental polarity is necessary to clarify whether the observed effects arise from user polarity or the pairing strategy.
%Consequently, this section addresses our central ambiguity by examining dialogue outcomes across systematically varied polarity levels. Specifically, we construct user profiles that span multiple degrees of polarity and investigate how the resulting dialogues' consistency and coherence evolve in each scenario.
In this section, we address this uncertainty by examining dialogue outcomes across systematically varied polarity levels. We construct user profiles that span multiple degrees of polarity and investigate how dialogue consistency and coherence evolve in each scenario.

\subsubsection{Experimental Setup}
\label{sec:rq1.2setup}

\paragraph{Defining Polarity Level}
We aim to determine whether dialogues between users with strongly polarized personas (e.g., clearly positive or negative) differ from those with weak or neutral emotional content. We define each persona’s polarity level based on the confidence score produced by the classification model used in \S\ref{sec:userprofile}: 
values closer to 0 indicate a highly negative polarity, those closer to 1 indicate a highly positive polarity, and scores near 0.5 represent a neutral persona.
%values closer to 0 imply highly negative polarity, those closer to 1 imply highly positive polarity, and those near 0.5 are classified as neutral. 
Because half of all personas lie at sentimental extremes, uniformly dividing confidence scores would yield imbalanced distributions across levels. Instead, we partition the score range into nine intervals that ensure a more balanced representation of persona instances. Further details on the partition criteria and the distribution across levels are provided in Appendix~\ref{sec:statistics}.

\paragraph{Experiment Pipeline}
We fix the number of personas $K$ to 1 in each user profile to minimize external factors such as the effect of combining or ordering multiple personas.
%The profile’s polarity level naturally follows the single persona’s level, and we pair profiles of the same level with each other. 
The profile's polarity level follows directly from that single persona, and we pair profiles of the same level together.
For each level, we create 500 distinct user pairs and generate dialogues using LLaMa-3.1-8B and Qwen-2.5-7B models. We then compute the average consistency and coherence scores—measured respectively by \textbf{C score} and \textbf{PairEval}, for each level.

\subsubsection{Results \& Analysis}
\label{sec:rq1.2results}

\begin{figure}[t]
    \centering
    \begin{subfigure}{0.47\textwidth}
        \centering
        \includegraphics[width=1\textwidth]{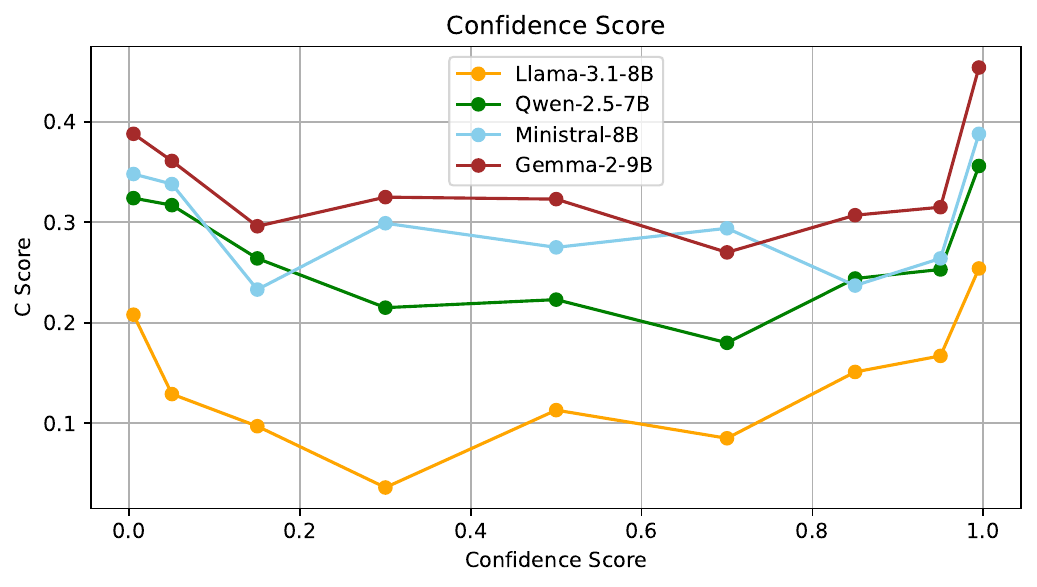}
        \caption{C score of per Confidence Score Interval}
        \label{fig:cs_cscore}
    \end{subfigure}
    
    \vspace{0.1cm} 
    
    \begin{subfigure}{0.47\textwidth}
        \centering
        \includegraphics[width=1\textwidth]{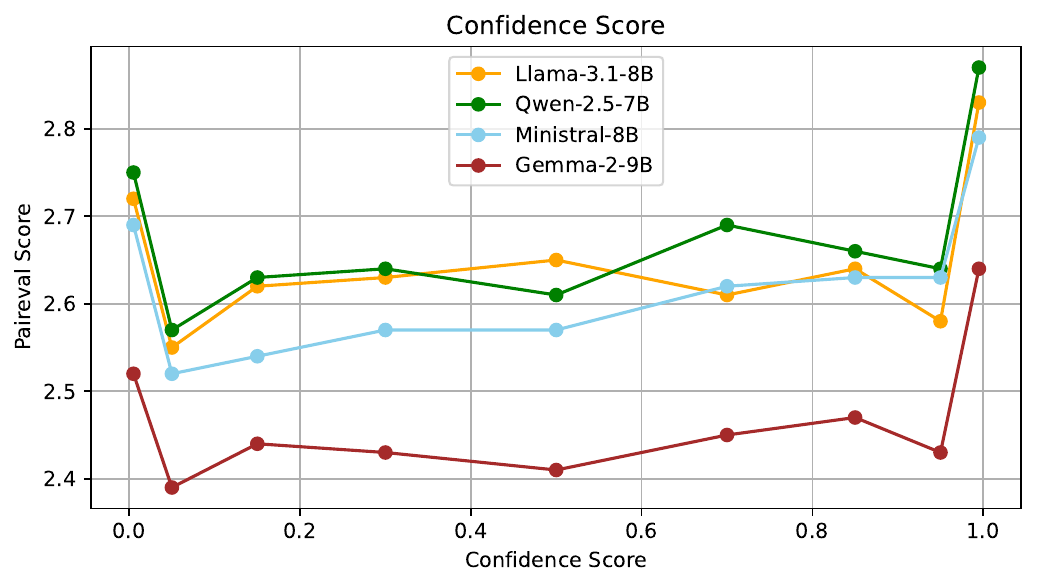}
        \caption{Paireval Score of per Confidence Score Interval}
        \label{fig:cs_paireval}
    \end{subfigure}
    
    %\caption{Consistency and Coherence Comparison Across Nine Polarity Levels. The dot's $x$-coordinate indicate the median confidence score for each level, where lower values (closer to 0) indicate more negative polarity and higher values (closer to 1) indicate more positive polarity.}
    \caption{Consistency and coherence across nine polarity levels. Dot $x$-coordinates show the median confidence score (0: most negative, 1: most positive).}
    \vspace{-2mm}
    \label{fig:rq1.2}
\end{figure}

\paragraph{Stronger Sentimental Extremity in Persona Yields Higher-Quality Dialogue}
As illustrated in Figure~\ref{fig:rq1.2}, both consistency and coherence scores exhibit a U-shaped trend with respect to polarity level. In other words, dialogues derived from highly polarized personas, whether strongly positive or strongly negative, tend to outperform those derived from more neutral personas. 
%This phenomenon appears in every tested LLM, suggesting that personalizing an LLM with a strongly polarized persona enhances dialogue quality.
This pattern is consistent across all tested models, suggesting that personalizing a model with a strongly polarized persona enhances dialogue quality.

\paragraph{Positive Polarity Results in Particularly High-Quality Dialogue}
Furthermore, dialogues generated from positively polarized profiles demonstrate superior consistency and coherence compared to those from negatively polarized profiles. Notably, the consistency score for dialogues produced by the LLaMa-3.1-8B model at the highest polarity level is up to seven times greater than that at the lowest level, while the Qwen-2.5-7B model shows an average improvement of approximately 0.3 points in coherence. These findings indicate that positive polarity yielding especially pronounced gains.

We further analyze the underlying causes of the performance gap discussed in this section in Appendix~\ref{sec:rq1_discussion}.

%\section{Does the polarity-based trend remain stable after applying factor scaling?
% Consistency in various experimental configurations
%Through Sec~\ref{sec:3, sec:4} we observed that as a user persona becomes more extreme—particularly closer to a positive polarity—the resulting dialogue quality tends to improve.
%Through Sec~\ref{sec:3}, we discussed why dialogues between negative users tend to yield higher \textbf{C score} and \textbf{Contd.} values yet remain relatively incoherent. By contrast, positive-user dialogues generally show lower Contd.\ values and higher coherence. Do these findings remain stable even after applying different factor scalings in our experimental design, or another intersting findings will be observed? This section addresses this question by scaling quantitative factors composing our expriments. In particular, we explore three quantitative factors: (1) The number of personas K in a user profile, (2) Polarity ratio of a user profile, and (3) Model size.

\section{How to Make LLMs Robust to Polarity?}
\label{sec:rq2}
%\vspace{-2mm}
%Our previous findings indicate that persona polarity influences dialogue quality in LLMs—dialogues generated from positive personas tend to be superior, while those with ambiguous or mixed polarity exhibit lower quality.
Our previous findings indicate that persona polarity influences dialogue quality in LLMs.
%This motivates our exploration of leveraging these traits to build a more robust personalized dialogue system. 
This observation underscores the need for methods that leverage polarity traits to build a more robust personalized dialogue system.
To that end, we conduct experiments using real-world profiles in an original pairing configuration to determine how to make LLMs robust to polarity.

\subsection{Adjusting Generation Strategy}
\label{sec:rq2.1}
In addition to the \textit{dual-persona joint generation} method, we adopt a \textit{turn-based agent conversation} approach (\S\ref{sec:design}). In this strategy, each turn is generated using a single user's profile, thereby avoiding the complexities associated with mixed polarity inputs (\S\ref{sec:rq1.1results}) and allows for better handling of persona-specific characteristics. 
%Although this approach requires a model call for each turn, our experiments in Appendix~\ref{sec:rq_appendix.3} indicate that even models with approximately 3B parameters can produce high-quality dialogues.
Although this approach requires a separate model call for each turn, our experiments show that models with around 3B parameters can generate high-quality dialogues without requiring extensive computational resources.
Consequently, we employ two LLMs, LLaMa-3.2-3B and Qwen-2.5-3B, in this configuration to reduce latency and computational load while maintaining performance.

When applying the turn-based approach, it is necessary to specify a dialogue endpoint. For a fair comparison, we align the number of turns with the distribution observed in the joint-generation method. For instance, if joint generation produces 10 dialogues—6 of which have 8-turns and 4 of which have 10—then we assign an 8-turn endpoint with a 60\% probability and a 10-turn endpoint with a 40\% probability before initiating turn-based dialogue generation.

%Moreover, generating utterances on a turn-by-turn basis sometimes results in excessively long responses, which cluttered the overall dialogue. To address this, we incorporated an instruction in the prompt to produce short and concise utterances. The detailed prompt is provided in the Appendix~\ref{sec:prompt}.
Additionally, generating utterances on a turn-by-turn basis sometimes results in excessively long responses that cluttered the overall dialogue. To address this, we add an instruction in the prompt to produce short and concise utterances. The detailed prompt is provided in Appendix~\ref{sec:prompt}.

\subsection{User Profile Ordering \& Sentiment-aware Prompting}
\label{sec:rq2.2}
Analysis in \S\ref{sec:rq1} revealed that profiles containing negative or neutral personas produce lower-quality dialogues compared to those with exclusively positive personas. Since modifying user personas is undesirable, we propose a simple but effective polarity-based ordering technique to enhance dialogue performance without altering the original content. 
Inspired by ~\cite{wu2024understanding} showing that LLMs perform better when desired persona content appears early in the prompt, we design three ordering strategies based on the confidence score from \S\ref{sec:userprofile}:
\begin{itemize}[leftmargin=5pt, itemsep=2pt, topsep=1pt, parsep=1pt]
    \item \textit{Score Ascending:} Arrange personas in order of increasing confidence (more negative first).
    \item \textit{Score Descending:} Arrange personas in order of decreasing confidence (more positive first).
    \item \textit{Center-Out Score Ascending:} Sort by $|(\text{confidence score} - 0.5)|$, adding a bias factor of $a=0.05$ for positive personas to prioritize negatives.
\end{itemize}
We apply this ordering only to the turn-based approach, as the joint generation method combines both profiles into a single prompt. 

Plus, we concatenate a lightweight prompt template: \textit{"Please ensure that each user's persona, especially negative or neutral personas, is well integrated into the dialogue and that the overall dialogue remains coherent"}. It selectively conditions the model on the sentiment of personas present in a user profile at the end of the generation prompt. Because large language models already handle positive personas reliably, the template explicitly instructs the model to be aware when responding to negative or neutral personas.

\subsection{Results \& Analysis}
\label{sec:rq2.3}

\begin{table}[h]
\centering
\setlength{\tabcolsep}{4pt}
\resizebox{0.49\textwidth}{!}{%
\begin{tabular}{l|l|ccc|cc}
\toprule
\multirow{2}{*}{\textbf{Model}} & \multirow{2}{*}{\textbf{Strategy}} 
& \multicolumn{3}{c|}{\textbf{Consistency}} 
& \multicolumn{2}{c}{\textbf{Coherence}} \\
 & & \textbf{C score} & \textbf{Contd.} & \textbf{G-eval} 
  & \textbf{PairEval} & \textbf{G-eval} \\
\midrule
\multirow{7}{*}{LLaMa-3.2} 
  & Joint 
   & 0.371 & 15.01 & 4.15
   & 2.70 & 4.50 \\
\cmidrule{2-7}
  & Turn-based 
   & 0.609 & 7.92  & 4.14
   & 2.79 & 4.56 \\
  & \quad + asc. 
   & 0.610 & 7.97  & 4.13
   & 2.78 & 4.65 \\
  & \quad + dsc. 
   & 0.597 & 8.09  & 4.13
   & 2.77 & 4.63 \\
  & \quad + c-asc.
   & 0.617 & 7.39  & 4.21
   & 2.79 & 4.67 \\
  & \quad + sap.
   & 0.688 & 6.56  & 4.18
   & 2.78 & 4.59 \\
  & \quad + c-sap.
   & \textbf{0.717} & \textbf{6.07}  & \textbf{4.25}
   & \textbf{2.84} & \textbf{4.68} \\
   
\midrule
\multirow{7}{*}{Qwen-2.5} 
  & Joint 
   & 0.470 & 11.68 & 4.32
   & 2.62 & 4.36 \\
\cmidrule{2-7}
  & Turn-based 
   & 0.557 & 10.45 & 4.02
   & 2.65 & 4.60 \\
  & \quad + asc.
   & 0.557 & 10.45 & 3.99
   & 2.69 & 4.69 \\
  & \quad + dsc.
   & 0.535 & 10.99 & 4.01
   & 2.67 & 4.69 \\
  & \quad + c-asc.
   & 0.570 & 10.07 & 4.08
   & 2.69 & 4.71 \\
  & \quad + sap.
   & \textbf{0.777} & 8.27  & 4.58
   & 2.61 & 4.63 \\
  & \quad + c-sap.
   & 0.774 & \textbf{7.49}  & \textbf{4.59}
   & \textbf{2.69} & \textbf{4.77} \\

\bottomrule
\end{tabular}
} % end \resizebox
%\caption{Comparison of dialogue consistency and coherence across different generation strategies. The experiments were conducted using backbone LLMs, Llama-3.2-3B and Qwen-2.5-7B, under the original pairing configuration. The abbreviations \textit{asc}, \textit{dsc}, and \textit{c-asc} denote the score ascending, score descending, and center-out score ascending techniques, respectively.}
\caption{Dialogue consistency and coherence across generation strategies using LLaMa-3.2-3B and Qwen-2.5-7B with the original pairing. Abbreviations: \textit{asc.} (score ascending), \textit{dsc.} (score descending), \textit{c-asc.} (center-out score ascending), \textit{sap.} (sentiment-aware prompting), \textit{c-sap.} (center-out score ascending + sentiment-aware prompting). Best results are in \textbf{bold}.}
\label{tab:rq2}
\vspace{-2mm}
\end{table}

\paragraph{Turn-based Approach Improves Overall Quality.}
\noindent
For the LLaMa-3.2-3B model, the \textit{turn-based} approach significantly enhances dialogue consistency compared to \textit{dual-persona joint generation}. This improvement stems from separating users’ polarized personas into distinct inputs, thereby reducing internal complexity. In contrast, the Qwen-2.5-3B model demonstrates lower \textbf{G-eval} scores primarily because it blends the personas of both users into a single dialogue. When \textbf{User 1} initiates the conversation, its persona tends to dominate, leading \textbf{User 2} to adopt a similar framing. For example, if \textbf{User 1}'s persona is \textit{"I enjoy sewing"}, the dialogue might begin with, \textit{"I heard you’re good at sewing"}, prompting \textbf{User 2} to continue discussing sewing. As a result, \textbf{User 2}'s responses inadvertently mix in \textbf{User 1}'s persona, while \textbf{User 1} seldom reaffirms the persona initially designated for \textbf{User 2}. This behavior ultimately lowers the overall consistency score. Nonetheless, contradictions between personas and utterances actually decrease, boosting contradiction metrics (\textbf{C score} and \textbf{Contd.}). Also, we can resolve this issue by prompt engineering, which can be further explored. 

%For the case of coherence, the overall score improves for both models, especially in Qwen-2.5-3B. This outcome indicates that sequentially processing an updated dialogue context supports better integration of previously generated content.
Regarding coherence, the overall score improves for both models, with a notable enhancement in the Qwen-2.5-3B model. This outcome suggests that sequentially processing an updated dialogue context supports better integration of previously generated content.

\paragraph{Additional Profile Ordering is Helpful.}
%As shown in Table~\ref{}, we show that among the three ordering strategies, the center-out ascending method yields modest yet consistent improvements across all five evaluation metrics.
As shown in Table~\ref{tab:rq2}, we show that the center-out ascending method yields modest yet consistent improvements across all five evaluation metrics compared to the other ordering strategies.
%In this approach, profiles are arranged so that those with moderate polarity appear first, while profiles with extreme negative traits are prioritized earlier in the sequence compared to clearly positive ones. 
In this approach, we arrange profiles so that those with moderate polarity appear first, while profiles with extreme negative traits are prioritized over clearly positive ones.
This ordering leverages the model's sensitivity to the arrangement of persona information~\cite{wu2024understanding}, enhancing personalization for ambiguous personas that are less effectively integrated (\S\ref{sec:rq1.2}) while enabling well-incorporated positive personas (\S\ref{sec:rq1}) to more robustly support the overall context. 
%Improved ordering heuristics and profile adjustment methods can be further explored.

\paragraph{Synergy between Profile Ordering and Sentiment-Aware Prompting}
%As shown in Table~\ref{}, we show that among the three ordering strategies, the center-out ascending method yields modest yet consistent improvements across all five evaluation metrics.
Furthermore, as shown in Table~\ref{tab:rq2}, adding the sentiment‑aware prompt alone produces some improvements for both models. More importantly, jointly applying the center‑out score‑ascending strategy and sentiment‑aware prompting yields a strong synergy that markedly elevates dialogue quality. In particular, the coherence metric increases by around 0.1 points for both models, demonstrating substantial gains in conversational coherence. We attribute this improvement to positioning the personas that require heightened attention earlier in the prompt, thereby capitalizing on the uni‑directional decoding of large language models in concert with the sentiment‑aware instruction.

\paragraph{Human Evaluation}
We further manually compare turn‑based generation with and without our profile ordering and sentiment‑aware prompting techniques.

\begin{table}[h]
\centering
\setlength{\tabcolsep}{4pt}
\resizebox{0.3\textwidth}{!}{%
\begin{tabular}{l|cc}
\toprule
\textbf{Strategy} & \textbf{Consistency} & \textbf{Coherence} \\
\midrule
Turn-based       & 1.80 & 2.40 \\
\quad + c-sap.          & 2.27 {\scriptsize\textcolor{red}{(+0.47)}} 
                        & 2.43 {\scriptsize\textcolor{red}{(+0.03)}} \\
\bottomrule
\end{tabular}
}
\caption{Human evaluation results for consistency and coherence of dialogues generated by the turn‑based approach, with and without profile ordering and sentiment‑aware prompting.}
%\vspace{-2mm}
\label{tab:rq2_human}
\end{table}

As shown in Table~\ref{tab:rq2_human}, under the original pairing condition, applying our profile ordering and sentiment‑aware prompting techniques to the turn‑based generation improves consistency by 0.47 points and coherence by 0.03 points compared with the baseline. These observations further reinforce the robustness and efficacy of our proposed approach.
 
\section{Related Works}
\label{sec:related}

\subsection{Personalized Dialogue Generation}
\label{sec:related.1} 
Personalized dialogue systems have increasingly focused on leveraging user-specific information for more contextually aligned interactions~\cite{li2016persona, zhang2018personalizing, zhang2019dialogpt, roller2020recipes}. Early approaches typically involved training generative models with VAE~\cite{lee2022improving} to ensure dialogue coherence or NLI components~\cite{song2021bob, chen2023learning, zhou2023simoap} to capture persona representations. These persona representations often defined as descriptive sentences~\cite{zhang2018personalizing, dinan2020second} or key-value attributes~\cite{qian2017assigning, gao2023livechat}. With the advent of large language models and subsequent instruction tuning methods~\cite{ouyang2022training}, persona information can now be more flexibly embedded directly into system prompts~\cite{yang2023palr, wang2023rolellm}.

Recent research has also advanced multi-agent~\cite{park2023generative} or dual-persona~\cite{xu2022cosplay, jandaghi2023faithful} strategies, enabling two distinct personas to converse within a single session. This approach enhances personalized-chatbot capabilities~\cite{shuster2022blenderbot, lee2023p5} and supports large-scale synthetic data generation~\cite{jandaghi2023faithful} for further personalization. 
%Our work aligns with these trends by examining how persona polarity—particularly positive, negative, and ambiguous sentiments—influences dialogue quality in LLM-driven systems.

\subsection{Dialogue Evaluation}
\label{sec:related.2} 
Evaluating open-domain dialogue is inherently multifaceted, reflecting diverse aspects such as coherence, fluency, and persona consistency~\cite{wang2024learning, samuel2024personagym}. Traditional metrics like ROUGE~\cite{lin2004rouge} and BLEU~\cite{papineni2002bleu} often fail to capture higher-level qualities. Consequently, specialized metrics leveraging pretrained models—including C score~\cite{madotto2019personalizing} (for consistency), QuantiDCE~\cite{ye2021towards}, and PairEval~\cite{park2024paireval} (for coherence)—have gained traction~\cite{ghazarian2022deam, li2024dialogues}. We adopt these automated metrics for quantifiable evaluation.

In parallel, LLM-based evaluation strategies have rapidly emerged as a cost-effective alternative to human annotation~\cite{chiang2023can}. Leveraging Chain-of-Thought prompting~\cite{wei2022chain} further enhances evaluative transparency, allowing models to articulate their reasoning~\cite{liu2023g}. In our work, we integrate both traditional and LLM-based metrics to comprehensively assess persona-driven dialogue.

\subsection{Sentimental Sensitivity in LLMs}
\label{sec:related.3} 
Large Language Models (LLMs) are known to be highly sensitive to a variety of factors, including prompt order, language, cultural context, and sentiment~\cite{lu2021fantastically, dang2024rlhf, shen2024understanding, kwok2024evaluating}. As a difference perspective from previous works, we focus on sentimental sensitivity in LLMs in our work.
Although instruction tuning~\cite{ouyang2022training} and RLHF~\cite{dang2024rlhf} can mitigate these effects, recent studies still show that contextual sentiment can strongly influence model outputs~\cite{liu2024large, wu2024evaluating}. However, while much of the existing research focuses on sentiment that arises naturally in generated text, less work has considered sentimental polarity embedded naturally in explicit persona definitions. This underexplored avenue is central to our investigation, as it can profoundly affect both the coherence and consistency of persona-driven dialogues.

\section{Conclusion}
In this work, we demonstrate that LLMs are sensitive to the sentimental polarity of the persona during generating personalized dialogues. While positively polarized users yield smooth interactions, negatively polarized users overemphasize persona attributes and lead to lower coherence and even consistency. Furthermore, we find that personas with weak or neutral sentiment generally produce lower-quality dialogues. This findings suggest that the quality of the dialogue decreases if the user profile contains negative, or even neutral personas. Consequently, we introduce a dialogue generation approach that leverages persona polarity through a turn-based strategy with profile ordering, achieving more consistent and coherent outputs. Our findings underscore the importance of incorporating persona sentiment into personalized dialogue systems.

\section*{Limitation}

Despite our extensive configurations designed to show that our observations are not confined to specific conditions, there remain areas for improvement. First, \textbf{the source dataset used in our study is limited.} To control for variations in LLM sensitivity that might arise from different domains or features of personas, we employed only a single dataset, focusing on sensitivity purely driven by sentimental polarity. Future work could expand upon our findings by employing alternative persona representations, including \textit{Sparse key-value attributes} or \textit{User history} in \S\ref{sec:dataset}. Nonetheless, the ConvAI2 dataset we used contains a substantial volume of data, offering sufficient permutations of persona configurations to validate our research.

Second, \textbf{the inherent bias of the backbone model.} Automated evaluation metrics may not align with human perception if the underlying models themselves are biased. Similarly, if the polarity classification model is biased, it could potentially misclassify a persona’s polarity based on certain keywords (e.g., those referencing race or culture). To mitigate such issues, we employ a broad set of diverse metrics to offset individual-model biases, and adopted confidence thresholds to classify polarity only when the model surpasses a predefined threshold (0.99). Furthermore, prior work indicates that BERT-based models exhibit fewer biases~\cite{wang2023primacy}, making complete inversions of our intuitive judgments relatively rare.

Finally, \textbf{context length vaires along the configurations}. Especially, negative polarity often involves negation words (e.g., “not,” “n’t”), potentially yielding longer sentences. However, Table~\ref{tab:statistics} indicates that the overall difference in length is minimal. Additionally, dialogues generated by LLaMa-3.2-3B using a turn-based approach are notably longer, complicating direct comparisons. Nonetheless, it is noteworthy that these more extended utterances maintain high coherence, suggesting a promising direction for dialogue quality and consistency.

\section*{Ethics Statement}
This research utilizes only fictional persona profiles from the ConvAI2 dataset, ensuring that the handling or storage of any real personal data is completely avoided. All dialogues generated by multiple large language models are entirely simulated; they represent virtual interactions created through algorithmic synthesis and not real human conversations. These simulated dialogues do not raise any privacy concerns, as they are devoid of actual personal data.

\section*{Acknowledgement}
This research was supported by the Chung-Ang University Graduate Research Scholarship in 2025. This work was supported by the Institute of Information \& Communications Technology Planning \& Evaluation (IITP) grant funded by the Korea government (MSIT) [RS-2021-II211341, Artificial Intelligence Graduate School Program (Chung-Ang University)].

\bibliography{acl_latex}
\clearpage
%This is an appendix.
\appendix

\section{Why Are Large Language Models Sensitive to Persona Sentiment?}
\label{sec:rq1_discussion}
We have observed the performance gap across different persona polarities in \S\ref{sec:rq1}. As we utilize an LLM without any additional fine-tuning, we could infer that LLMs inherently exhibit sensitivity to persona sentiment. Consequently, we explain that this sensitivity is mainly due to the \textbf{scarcity of negative expressions in the pretraining data}, which may have been intentionally limited to prevent toxicity. And Figure~\ref{fig:per_stat} clearly illustrates a lower proportion of negative personas compared to other polarities. Such data imbalance is evident even in widely-used foundational datasets like ConvAI2, suggesting similar limitations in collecting sufficient negative persona data in other datasets or real-world contexts. Additionally, we also find that LLM’s sensitivity can stem from post-training techniques that steer models toward generating responses preferred by humans. For the experiment, we sample 1,000 dialogues per model under both the negative and positive configurations described in \S\ref{sec:rq1.1setup}. For the neutral configuration, we follow a similar approach by grouping five intermediate-stage personas from Figure~\ref{fig:rq1.2} into pairs, generating 1,000 dialogues for each model accordingly. 

\begin{table}[h]
  \centering
  \resizebox{0.3\textwidth}{!}{%
  \begin{tabular}{l|c|c}
    \toprule
    \textbf{Pairing} & \textbf{Persona} & \textbf{Dialogue} \\
    \midrule
    Negative & 1.71 & 2.67 {\small\textcolor{red}{(+0.96)}} \\
    Positive & 4.25 & 4.52 {\small\textcolor{red}{(+0.27)}} \\
    Neutral  & 3.06 & 4.09 {\small\textcolor{red}{(+1.03)}} \\
    \bottomrule
  \end{tabular}
  }
  \caption{Average sentiment scores (1–5 Likert) for dialogues and their personas, rated by GPT‑4o. Lower scores denote negativity, higher positivity; red values mark the dialogue’s sentiment gain over its persona.}
  \label{tab:discuss}
  \vspace{-3mm}
\end{table}

As shown in Table~\ref{tab:discuss}, regardless of model type, we observe a consistent tendency to produce more positive dialogues than the assigned personas. Models consistently generate more positive dialogues, with the gap widening when dialogue quality decreases. This trend suggests that \textbf{post-training practices bias models toward positive outputs, leading to confusion} when persona sentiment diverges significantly, lowering dialogue quality.

\section{Which Factors Influence the Polarity-Based Trend?}
\label{sec:rq_appendix}
In \S\ref{sec:rq1}, we observe that dialogues between negative users tend to yield higher \textbf{C score} and \textbf{Contd.} values while remaining relatively incoherent, whereas dialogues from positive users generally exhibit lower \textbf{Contd.} values and improved coherence. This section investigates whether these trends persist under variations in key experimental factors or if new patterns emerge. Specifically, we examine the influence of: (1) the number of personas \(K\) in a user profile, (2) the polarity ratio within a user profile, and (3) the model size.

\subsection{Varying Persona Count}
\label{sec:rq_appendix.1}
\paragraph{Experiment Pipeline}
We vary the value of number of personas \(K\) in each user profile to 1, 2, and 10, then examine how these changes affect consistency and coherence in dialogues. Adopting the method from \S\ref{sec:userprofile}, we generate a total of 1K user profiles per polarity and create 1K user pairs according to the \textit{Negative}, \textit{Positive}, and \textit{Mixed Pairing} strategies described in \S\ref{sec:rq1}. When each profile contains only one persona, \textit{Mixed Pairing} is not feasible; we therefore use \textit{Opposite Pairing} in that case. We measure consistency with the \textbf{C score} and \textbf{Contd.}\ and coherence with the \textbf{PairEval} score.

\begin{figure}[h]
    \centering
    \begin{subfigure}{0.47\textwidth}
        \centering
        \includegraphics[width=1\textwidth]{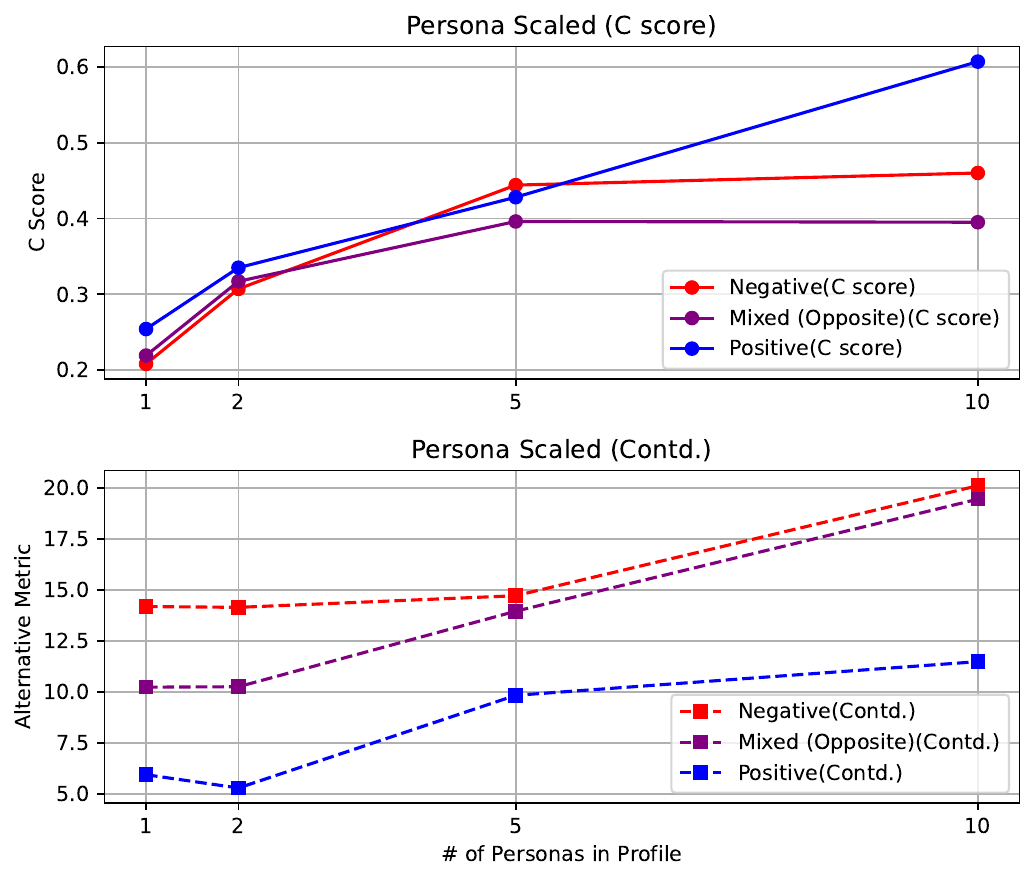}
        \caption{C score and Contd.\ for Scaled K Values. The solid lines represent the trend in \textbf{C score}, while the dotted lines indicate the trend in \textbf{Contd.}}
        \label{fig:ps_cscore}
    \end{subfigure}
    
    \vspace{0.1cm} 
    
    \begin{subfigure}{0.47\textwidth}
        \centering
        \includegraphics[width=1\textwidth]{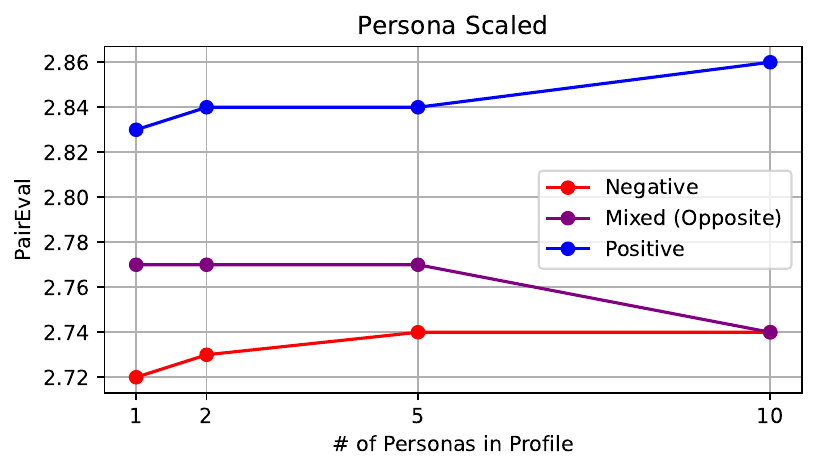}
        \caption{Paireval Score for Scaled K values.}
        \label{fig:ps_paireval}
    \end{subfigure}
    
    \caption{Consistency and Coherence Comparison Between Pairing Recipe. K is set to 1, 2, 5, and 10. We use Llama-3.1-8B for backbone LLM.}
    \label{fig:ps}
    \vspace{-5mm}
\end{figure}

\paragraph{\textit{Positive Pairing} Exhibits a Direct Positive Relationship between K and Dialogue Quality}
As shown in Figure~\ref{fig:ps}, dialogues generated under \textit{Positive pairing} achieve the highest consistency and coherence scores except for C score in K=5 setting. Specifically, Figure~\ref{fig:ps_cscore} reveals a positive correlation between K and the \textbf{C score} for \textit{Positive Pairing}, despite of growing \textbf{Contd.}\ values. It suggests that as the number of positive personas grows, LLMs can still effectively incorporate them while suppressing the increase in the number of contradictions. Interestingly, as illustrated in Figure~\ref{fig:ps_paireval}, coherence also improves with more personas in \textit{Positive Pairing}, contrary to our initial assumption that additional personas might artificially distort the conversation context.

\paragraph{\textit{Negative} and \textit{Mixed Pairing} Show No Clear Improvement}
\noindent
In contrast, \textit{Negative} and \textit{Mixed Pairings} exhibit only marginal gains in consistency and coherence as $K$ increases. Notably, when $K$ is either small or very large, contradictions accumulate faster than entailments, causing the C score to drop below that of positive pairings. In particular, \textit{Mixed Pairing} displays a declining trend in both metrics, presumably because the model struggles to integrate multiple personas polarized at opposite extremes.

%\subsection{Scaling the Polarity Ratio of User Profile}
\subsection{Impact of Polarity Ratio}
\label{sec:rq_appendix.2}

\begin{figure}[h]
    \centering
    \includegraphics[width=0.45\textwidth]{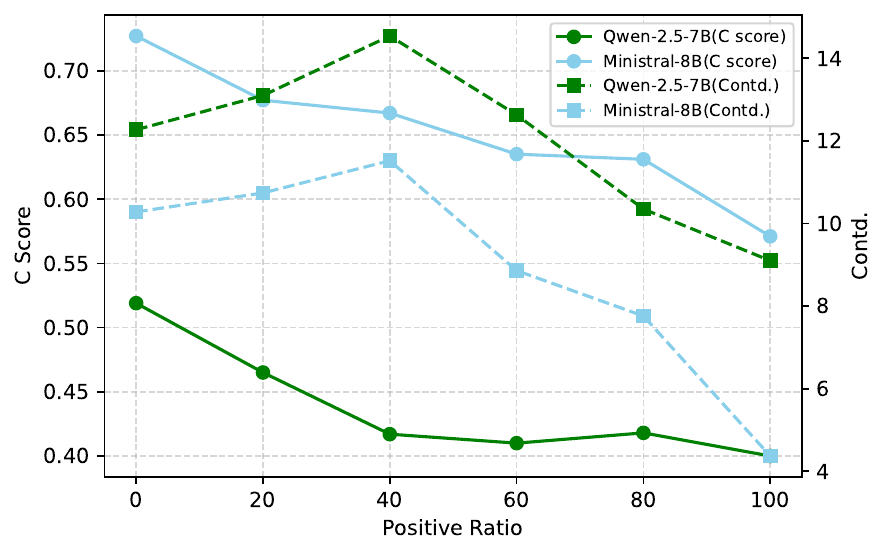}
    %\caption{C Score by the Proportion of Positive Personas in Each User Profile. We use LLaMa-3.1-8B and Ministral-8B as the backbone LLMs. As the positive ratio increases, the consistency of the generated dialogues tends to improve.}
    \caption{C Score by Positive Persona Ratio. Using LLaMa-3.1-8B and Ministral-8B, higher positive ratios correlate with improved dialogue consistency.}
    \vspace{-5mm}
    \label{fig:rs}
\end{figure}

\paragraph{Experiment Pipeline}
Next, we fix K to 5 and vary only the proportion of positive personas to assess its influence on dialogue generation. Building on the \textit{Mixed Pairing} setup in \S\ref{sec:rq1}, we categorize user profiles by their positive-persona ratios and measure the consistency of the resulting dialogues with the \textbf{C score} and \textbf{Contd.} Since each paired user may have a different proportion of positive personas, we calculate consistency of the dialogue separately for each user, then average the scores across each ratio category. We use two backbone LLMs, Llama-3.1-8B and Ministral-8B, which yield highly consistent dialogue in \S\ref{sec:rq1.1}.

\paragraph{A Notable Anomaly in Mixed Ratios}
As shown in Figure~\ref{fig:rs}, the \textbf{C score} decreases as the proportion of positive personas grows, in line with our initial expectations. However, \textbf{Contd.}\ does not steadily decline; it first rises and then starts to drop once the positive ratio exceeds 50\%. This pattern consistently appears in both models, implying that when a profile contains negative personas mixed with a slightly smaller share of positive personas, the model becomes more overemphasized to incorporate the user persona. Nonetheless, because the distribution of data across different positive ratios is uneven and our sample size is relatively small, these findings serve only as preliminary observations rather than definitive conclusions.

\subsection{Effect of Model Size}
\label{sec:rq_appendix.3}

\paragraph{Experiment Pipeline}
To determine whether polarity-based trends remain consistent across different model sizes, we employ various Qwen-2.5 models ranging from 0.5B to 3B, 7B, 14B, and 32B, alongside the default 7B model. For practical purposes, we sample 1K user pairs from the pool using the \textit{Negative Pairing}, \textit{Positive Pairing}, and \textit{Mixed Pairing} strategies in \S\ref{sec:rq1}, ensuring identical inputs for each model. The dialogue metrics follow those in Appendix\ref{sec:rq_appendix.1}.

\begin{figure}[h]
    \centering
    \begin{subfigure}{0.5\textwidth}
        \centering
        \includegraphics[width=1\textwidth]{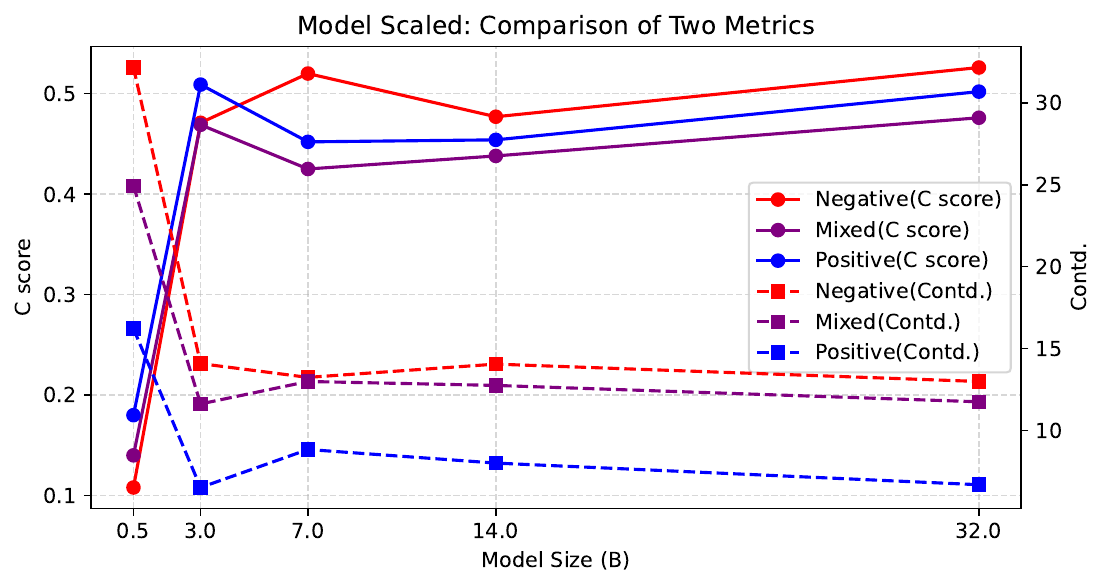}
        \caption{C score and Contd.\ for Different Model Sizes. The solid lines represent the trend in C score, while the dotted lines indicate the trend in Contd.}
        \label{fig:ms_con}
    \end{subfigure}
    
    \vspace{0.1cm} 
    
    \begin{subfigure}{0.47\textwidth}
        \centering
        \includegraphics[width=1\textwidth]{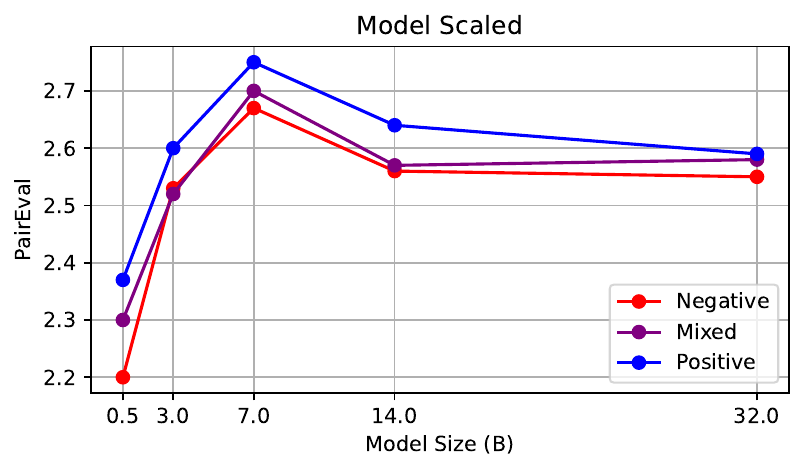}
        \caption{Paireval Score for Scaling the Model Size.}
        \label{fig:ms_coh}
    \end{subfigure}
    
    \caption{Consistency and Coherence Comparison Between Model Size. We use Qwen-2.5-0.5B, 3B, 7B, 14B and 32B for various scaled models.}
    \label{fig:ms}
    \vspace{-5mm}
\end{figure}

\paragraph{Polarity-Based Trends Fluctuate with Model Size}
In \S\ref{sec:rq1.1}, we discussed why dialogues between negative users tend to yield higher \textbf{C score} and \textbf{Contd.} values yet remain relatively incoherent. By contrast, positive-user dialogues generally show lower Contd.\ values and higher coherence. As illustrated in Figure~\ref{fig:ms_con}, these tendencies shift with model size. When the model is 3B or smaller (i.e., 0.5B or 3B), negative-user dialogues exhibit markedly low \textbf{C score} and high \textbf{Contd.}, indicating that smaller-scale models handle negative personas less effectively than positive personas. However, once the model size surpasses 7B (e.g., 7B, 14B, 32B), the original pattern reemerges, suggesting that while larger models gain the capacity to process negative personas more accurately, they continue to overemphasize them in the dialogue.

\paragraph{Dialogue Quality Does Not Always Increase with Model Size}
We initially hypothesized that a larger model would generate higher consistency and coherence; however, the experimental results do not fully support this assumption. For consistency, the 3B model actually achieves the highest average \textbf{C score} and \textbf{Contd.}, irrespective of user polarity. Although performance improves somewhat as model size scales from 7B to 32B, the gains are not substantial relative to the increase in size. In terms of coherence, the performance peaks at 7B and then declines at larger scales. These findings align with ~\cite{chen2023bigger}'s conclusion that model size does not necessarily translate into higher downstream task performance.

\section{GPT-4o Results}
\label{sec:gpt}
We extended our analysis by incorporating GPT‑4o—the leading closed‑source LLM in many chatbot benchmarks—into our backbone set to test whether the phenomenon identified in \textbf{RQ1} persists. The experimental setup mirrors that described in \S\ref{sec:rq1.1setup}.

\begin{table}[h]
\centering
\setlength{\tabcolsep}{4pt}
\resizebox{0.48\textwidth}{!}{%
\begin{tabular}{l|ccc|cc}
\toprule
\multirow{2}{*}{\textbf{Pairing}} 
& \multicolumn{3}{c|}{\textbf{Consistency}} 
& \multicolumn{2}{c}{\textbf{Coherence}} \\
\cmidrule(lr){2-4} \cmidrule(lr){5-6}
& \textbf{C score} \(\uparrow\) & \textbf{Contd.} \(\downarrow\) & \textbf{G-eval} \(\uparrow\)
& \textbf{PairEval} \(\uparrow\) & \textbf{G-eval} \(\uparrow\) \\
\midrule
\textit{Original} & 0.385 & 7.92 & 4.14 & 2.79 & 4.56 \\
\textit{Negative} & 0.626 & 11.11 & 4.27 & 2.75 & 4.44 \\ 
\textit{Positive} & \textbf{0.651} & \textbf{4.12} & \textbf{4.42} & \textbf{2.83} & \textbf{4.69} \\
\textit{Mixed}    & 0.638 & 8.60 & 4.32 & 2.76 & 4.54 \\
\textit{Opposite} & 0.589 & 8.64 & 4.23 & 2.75 & 4.44 \\
\bottomrule
\end{tabular}
} % end of resizebox
\caption{Combined consistency and coherence results of the dialogues generated by GPT-4o.}
\vspace{-2mm}
\label{tab:gpt}
\end{table}

As shown in Table~\ref{tab:gpt}, the same pattern emerges with GPT‑4o: the \emph{positive‑pairing} configuration yields the highest dialogue quality across all metrics, whereas configurations containing negative pairings or mixed sentiment polarities perform noticeably worse. In particular, the \emph{original‑pairing} configuration—which reflects real‑world distributions—suffers a larger quality drop than we observed with open‑source LLMs. This result reinforces the robustness of our findings and warns of potential, unforeseen performance degradation in downstream applications.

\section{Evaluation}
\label{sec:detailed_evaluation}
\subsection{Consistency}
\label{sec:detailed_consistency}
\paragraph{C score}
\noindent
\textbf{C score}~\cite{madotto2019personalizing} is a widely used metric for assessing the consistency of personalized utterances. Let the user profile for user \(n\) be \(U^{(n)}\) and its constituent personas be \(p_i^{(n)}\). Similarly, denote the dialogue by \(D\), composed of utterances \(u_i^{(n)}\) from user \(n\). We feed these persona-utterance pairs into a fine-tuned NLI model to obtain entailment scores, then sum these values for each utterance across all personas:
\[
U^{(n)} = \{\,p_1^{(n)},\,p_2^{(n)},\,\dots,\,p_k^{(n)}\}\quad,
\]
\[
D = \{\,u_1^{(1)},\; u_1^{(2)},\;u_2^{(1)},\;u_2^{(2)},\;\dots\}
\]
\[
\mathrm{NLI}\bigl(u_i^{(n)},\,p_j^{(n)}\bigr) \;=\;
\begin{cases}
  1 & \text{entailment},\\
  0 & \text{neutral},\\
  -1 & \text{contradiction}.
\end{cases}
\]

\[
C\bigl(u_i^{(n)}\bigr) \;=\; \sum_{j=1}^k 
  \mathrm{NLI}\bigl(u_i^{(n)},\;p_j^{(n)}\bigr)
\]

We fine-tune a BERT-large~\cite{devlin2018bert} model on the DialogNLI dataset~\cite{welleck2018dialogue} by utilizing AdamW optimizer~\cite{loshchilov2017decoupled} for NLI, achieving near-state-of-the-art accuracy (88.9\%) on the test set, using a learning rate of \(1\times 10^{-5}\) and a batch size of 16.

Originally, the C score was designed to measure the consistency of a single response given a dialogue context. To adapt it for entire dialogues, we split the dialogue into individual utterances, match each utterance of speaker \(n\) to its user profile \(U^{(n)}\), compute all persona-utterance entailment scores, and sum them. We then average these sums across all utterances to obtain the dialogue-level C score:
\[
C(D) 
= \frac{1}{\lvert D\rvert}
  \sum_{i}
  \Bigl(
    C\bigl(u_i^{(1)}\bigr)
    \;+\;
    C\bigl(u_i^{(2)}\bigr)
  \Bigr),
\]
where \(\lvert D\rvert\) is the total number of turns in the dialogue.

\paragraph{Contd.}
\noindent
Because the C score alone may be susceptible to Simpson’s Paradox~\cite{simpson1951interpretation}, we propose the Contradiction Ratio \textbf{(Contd.)} to provide a more detailed distribution of entailments versus contradictions. For example, suppose the entailment scores for a particular utterance across personas are \([0, 0, 0, 1, 0]\) and \([1, 1, 1, -1, 0]\). These yield C scores of 1 and 2, respectively, even though the latter contains a contradiction. Consequently, Contd.\ measures the percentage of contradictions out of all entailments or contradictions. Formally:
\[
\begin{alignedat}{1}  % ← 'alignedat{1}'로 불필요한 들여쓰기 방지
\mathrm{\#\; of\; Contradictions\; (C\#)} = \\
\sum_{n=1}^{2} \sum_{i} \sum_{j=1}^{k}
  \mathbf{1}\Bigl(\mathrm{NLI}(u_i^{(n)},\,p_j^{(n)}) = -1\Bigr),\\
\mathrm{\#\; of\; Entailments\; (E\#)} = \\
\sum_{n=1}^{2} \sum_{i} \sum_{j=1}^{k}
  \mathbf{1}\Bigl(\mathrm{NLI}(u_i^{(n)},\,p_j^{(n)}) = 1\Bigr).
\end{alignedat}
\]

\[
\mathrm{Contd.} 
= \left(
    \frac{\mathrm{C\#}}
         {\mathrm{C\#} + \mathrm{E\#}}
  \right) \times 100,
\]
where \(\mathbf{1}(\cdot)\) is the indicator function.

\subsection{Coherence}
\label{sec:detailed_coherence}
\paragraph{Q-DCE \& PairEval}
\noindent
Traditional reference-based metrics (e.g., BLEU~\cite{papineni2002bleu}, METEOR~\cite{banerjee2005meteor}) often fail to correlate well with human judgments of dialogue coherence~\cite{ghazarian2022deam, li2024dialogues}, inspiring the emergence of reference-free automatic evaluation. Among these, Q-DCE uses a ranking-based loss function to pre-train an encoder model~\cite{devlin2018bert}, then applies knowledge distillation to produce quantitative coherence scores on a scale of 1 to 5.

With the rise of powerful LLMs, reference-free evaluation has increasingly shifted toward large language models~\cite{liusie2023zero}. We adopt \textit{PairEval}~\cite{park2024paireval}, wherein a fine-tuned LLM compares a target response against alternative responses given the same dialogue context. However, these studies often focus on single-turn evaluations, so we treat every utterance from the second to the last as a candidate response. All preceding utterances are considered the dialogue context for that response. Formally, let \(D = \{u_1, u_2, \dots, u_N\}\) represent the dialogue, and let \(M(c, u) = s\) denote an automated evaluation model that produces a coherence score \(s\) for a response \(u\) given context \(c\). We define the coherence of the dialogue \(\mathrm{Coh}(D)\) as:
\[
\mathrm{Coh}(D) 
= \frac{1}{N - 1} 
  \sum_{i=2}^{N}
  M\bigl(u_{1:i-1},\,u_i\bigr).
\]

\section{Prompt Templates}
\label{sec:prompt}
\begin{figure*}[!htbp]
    \centering
    \includegraphics[width=0.9\textwidth]{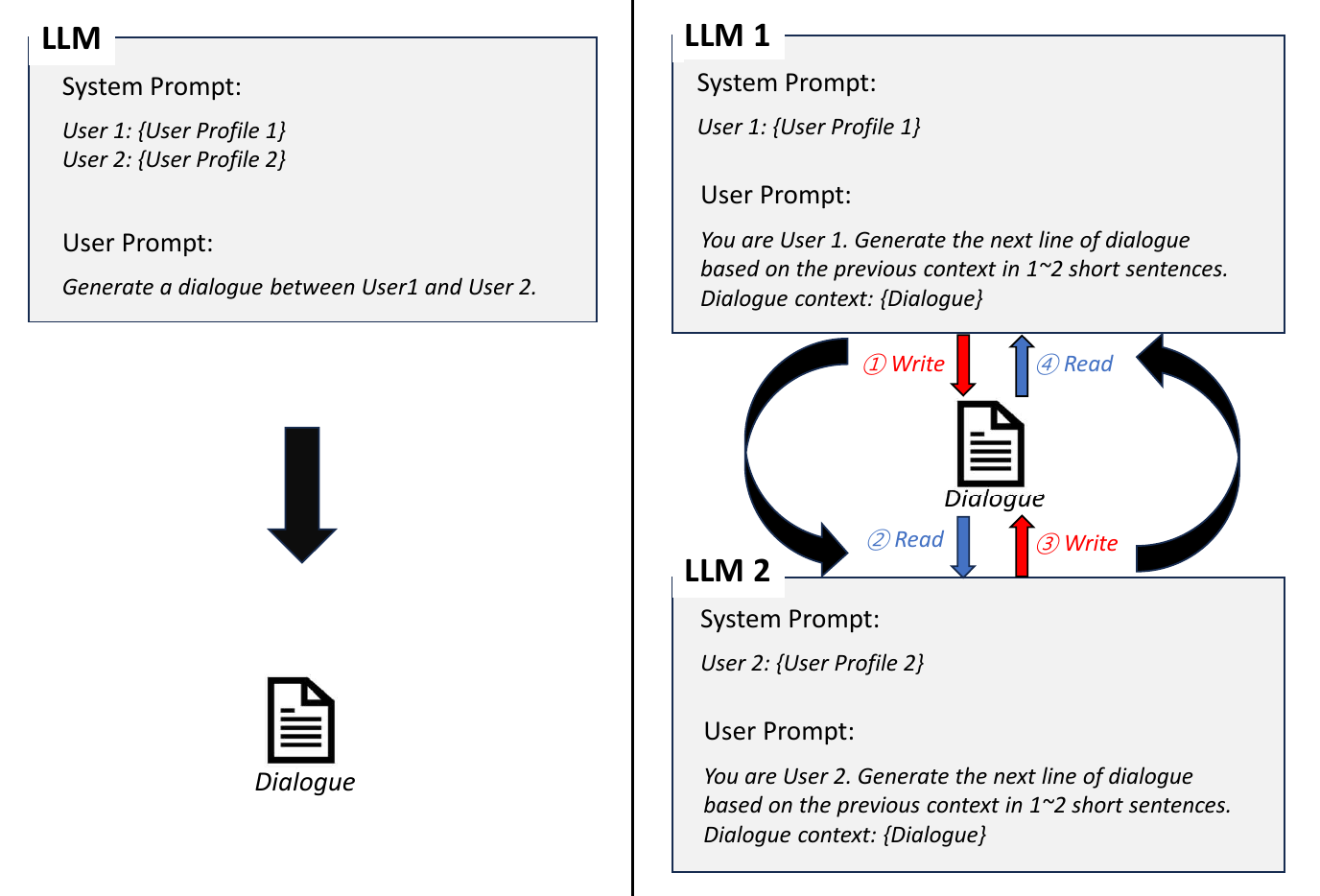}
    %\caption{C Score by the Proportion of Positive Personas in Each User Profile. We use LLaMa-3.1-8B and Ministral-8B as the backbone LLMs. As the positive ratio increases, the consistency of the generated dialogues tends to improve.}
    \caption{Prompt Template with Simple Framework Illustration for Two Generation Strategies. First one indicates \textit{dual-persona joint generation}, and second one indicates \textit{turn-based approach}.}
    \label{fig:Generation}
\end{figure*}

In Figure~\ref{fig:Generation}, we present the prompt templates for the two generation strategies introduced in \S\ref{sec:generation} and employed in \S\ref{sec:rq2}. Left part shows the LLM prompt used in \S\ref{sec:rq1} and \S\ref{sec:rq2} whereas right part illustrates the LLM prompt applied in \S\ref{sec:rq2}. In the latter, two LLMs are personalized with distinct user profiles, and additional instructions are included to limit the length of the responses.

\begin{figure*}[htbp]
    \centering
    % 첫 번째 서브 피규어
    \begin{subfigure}[b]{0.9\linewidth}
        \centering
        \includegraphics[width=\linewidth]{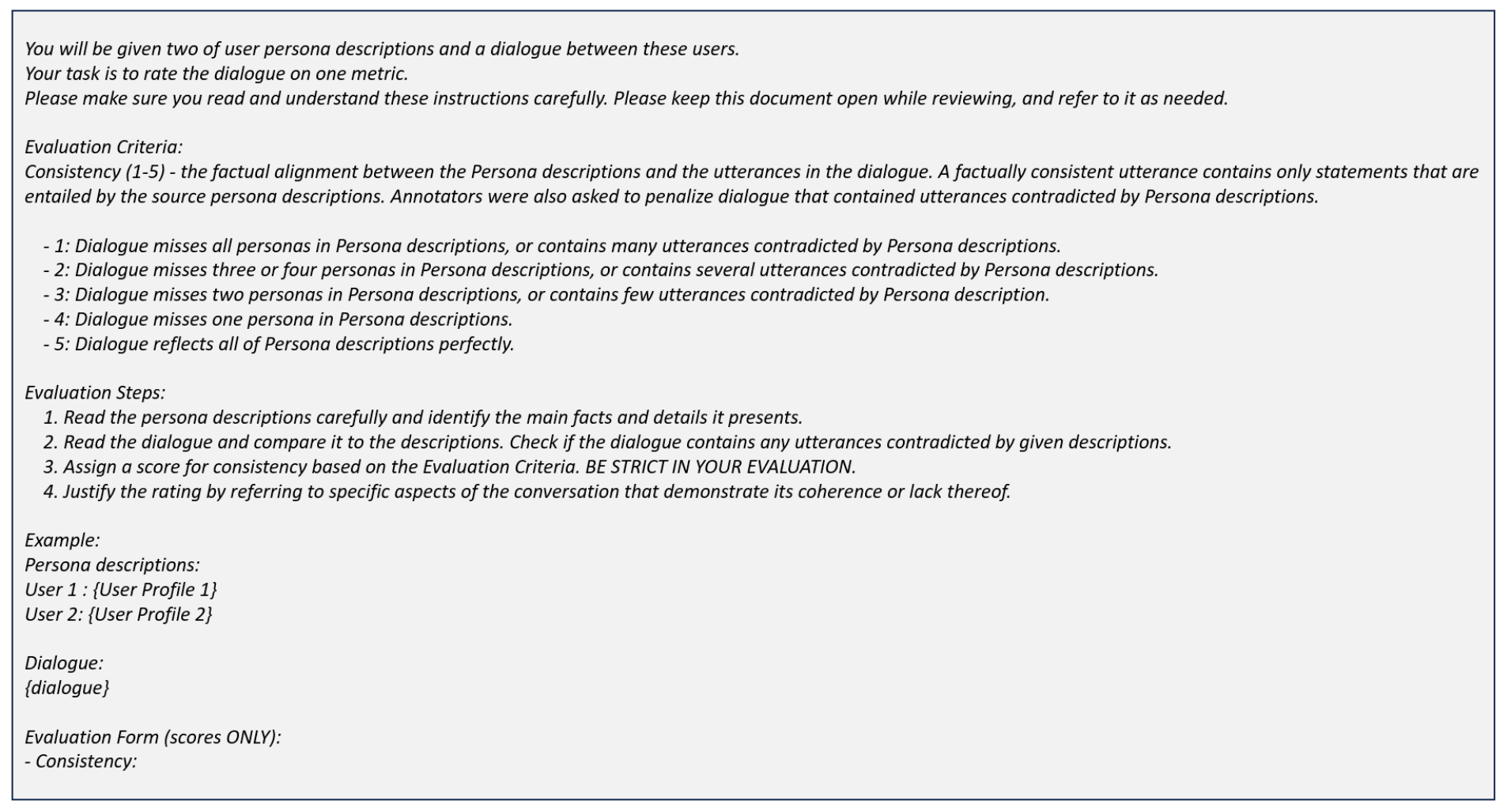} % 그림 파일명 입력
        \caption{G-eval Prompt for Evaluating Consistency.}
        \label{fig:G-eval1}
    \end{subfigure}
    \vspace{1em} % 그림 사이의 수직 간격 조정
    % 두 번째 서브 피규어
    \begin{subfigure}[b]{0.9\linewidth}
        \centering
        \includegraphics[width=\linewidth]{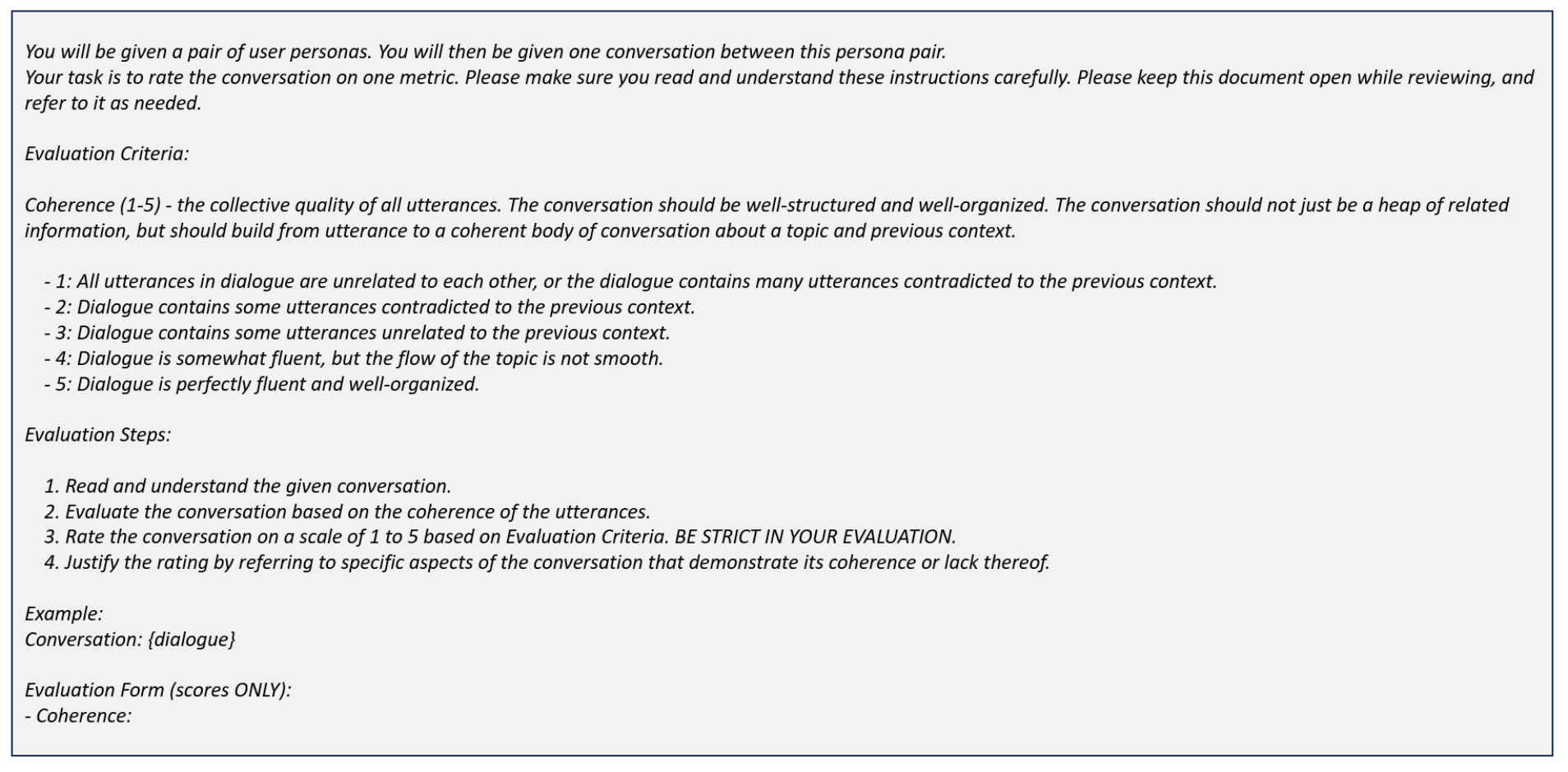} % 그림 파일명 입력
        \caption{G-eval Prompt for Evaluating Coherence}
        \label{fig:G-eval2}
    \end{subfigure}
    \caption{G-eval Prompts for Evaluating Dialogue Quality.}
    \label{fig:G-eval}
\end{figure*}

Furthermore, Figure~\ref{fig:G-eval} depicts the GPT evaluation prompt template introduced in \S\ref{sec:evaluation}. Both templates provide detailed scoring criteria to facilitate more accurate evaluations by the models. Since it was common for dialogues to receive a perfect score of 5, we added the instruction “BE STRICT TO YOUR EVALUATION” to promote a wider distribution of scores. Additionally, to distinguish our approach from conventional metrics, a stronger penalty for contradictions was imposed.

\section{Dialogue Statistics}
\label{sec:statistics}
\begin{table*}[ht!]
\centering
\small
\setlength{\tabcolsep}{8pt} % Tighten column spacing if needed
\resizebox{0.97\textwidth}{!}{%
\begin{tabular}{l l r r r r}
\toprule
\textbf{Source} & \textbf{Personas (profiles)} & \textbf{\# of samples} & \textbf{\# profile words} & \textbf{\# dialogue turns} & \textbf{\# dialogue words}\\
\midrule

%----------------- CONVAI2 (11 rows) -----------------
\multirow{11}{*}{\textbf{ConvAI2}} 
 & unique profiles & 2310 & 63.34 & - & - \\
 & unique personas & 6126 & 6.59 & - & - \\
 & negative personas (confidence score $\leq 0.01$) & 1006 & 7.08 & - & - \\
 & $0.01 < \text{confidence} \leq 0.1$ & 652 & 6.88 & - & - \\
 & $0.1 < \text{confidence} \leq 0.2$ & 158 & 7.02 & - & - \\
 & $0.2 < \text{confidence} \leq 0.4$ & 151 & 6.53 & - & - \\
 & $0.4 < \text{confidence} \leq 0.6$ & 150 & 6.63 & - & - \\
 & $0.6 < \text{confidence} \leq 0.8$ & 194 & 6.47 & - & - \\
 & $0.8 < \text{confidence} \leq 0.9$ & 191 & 6.76 & - & - \\
 & $0.9 < \text{confidence} \leq 0.99$ & 933 & 6.57 & - & - \\
 & positive personas (confidence score $> 0.99$) & 2691 & 6.42 & - & - \\
\midrule

%----------- Synthesized user profiles (3 rows) -------
\multirow{3}{*}{\shortstack[l]{\textbf{Synthesized}\\ \textbf{user profiles}}}
 & Negative Profiles & 10000 & 35.68 & - & - \\
 & Mixed Profiles    & 10000 & 34.65 & - & - \\
 & Profiles Profiles & 10000 & 35.07 & - & - \\
\midrule

%-------------------- RQ1 (29 rows) -------------------
\multirow{20}{*}{\S\ref{sec:rq1.1}}
 & Original pairings (llama) & 2296 & 63.27 & 16.01 & 414.09 \\
 & Negative pairings (llama) & 2898 & 70.26 & 14.53 & 390.15 \\
 & Mixed pairings (llama)    & 2879 & 69.34 & 15.36 & 417.49 \\
 & Positive pairings (llama) & 2864 & 67.29 & 14.86 & 423.67 \\
 & Opposite pairings (llama) & 2826 & 69.47 & 15.82 & 411.73 \\
 & Original pairings (qwen)  & 2301 & `` & 10.42 & 356.15 \\
 & Negative pairings (qwen)  & 2987 & `` & 10.05 & 393.20 \\
 & Mixed pairings (qwen)     & 2916 & `` & 10.43 & 390.12 \\
 & Positive pairings (qwen)  & 2851 & `` & 10.46 & 374.32 \\
 & Opposite pairings (qwen)  & 2895 & `` & 10.52 & 381.12 \\
 & Original pairings (ministral) & 2288 & `` & 10.73 & 287.17 \\
 & Negative pairings (ministral) & 2940 & `` & 9.88  & 289.77 \\
 & Mixed pairings (ministral)    & 2851 & `` & 10.10 & 295.07 \\
 & Positive pairings (ministral) & 2790 & `` & 10.13 & 299.75 \\
 & Opposite pairings (ministral) & 2801 & `` & 10.52 & 299.20 \\
 & Original pairings (gamma)     & 2167 & `` & 13.66 & 227.85 \\
 & Negative pairings (gamma)     & 2710 & `` & 13.43 & 233.62 \\
 & Mixed pairings (gamma)        & 2795 & `` & 13.97 & 239.88 \\
 & Positive pairings (gamma)     & 2863 & `` & 12.72 & 230.19 \\
 & Opposite pairings (gamma)     & 2672 & `` & 15.12 & 236.88 \\
 & Negative pairings (K=1) (llama) & 500 & 13.61 & 10.29 & 281.76 \\
 \midrule
 \multirow{9}{*}{\S\ref{sec:rq1.2}}
 & ($0.01 < \text{conf} \leq 0.1$) pairings (llama)  & 500 & 13.64 & 11.46 & 289.12 \\
 & ($0.1 < \text{conf} \leq 0.2$) pairings (llama)   & 500 & 14.36 & 11.40 & 275.91 \\
 & ($0.2 < \text{conf} \leq 0.4$) pairings (llama)   & 500 & 13.41 & 11.27 & 228.41 \\
 & ($0.4 < \text{conf} \leq 0.6$) pairings (llama)   & 500 & 12.84 & 10.95 & 263.11 \\
 & ($0.6 < \text{conf} \leq 0.8$) pairings (llama)   & 500 & 13.49 & 11.45 & 231.23 \\
 & ($0.8 < \text{conf} \leq 0.9$) pairings (llama)   & 500 & 13.02 & 11.27 & 253.59 \\
 & ($0.9 < \text{conf} \leq 0.99$) pairings (llama)  & 500 & 12.96 & 12.64 & 241.78 \\
 & Positive pairings (K=1) (llama) & 500 & 12.87 & 9.72 & 284.83 \\
\midrule
%-------------------- RQ2 (10 rows) -------------------
\multirow{10}{*}{\S\ref{sec:rq2}}
 & Original pairing (llama-3B) & 2203 & 63.26 & 10.33 & 231.17 \\
 & Original pairing (qwen-3B)  &  2254  & ``    & 11.55 & 246.13 \\
 & Original pairing (llama-turn-based) & 2032 & `` & 17.68 & 1020.30 \\
 & Original pairing (qwen-turn-based)  & 2190 & `` & 12.03 & 382.63 \\
 & Original pairing (llama-turn-based + ascending)   & 2056 & `` & 17.56 & 1048.87 \\
 & Original pairing (llama-turn-based + descending)  & 2060 & `` & 17.86  & 1056.91 \\
 & Original pairing (llama-turn-based + center-out Ascending) & 2060 & `` & 17.43 & 1031.42 \\
 & Original pairing (qwen-turn-based + ascending)    &  & 2148 & 12.09 & 256.13 \\
 & Original pairing (qwen-turn-based + descending)   & 2181 & `` & 12.09 & 266.45 \\
 & Original pairing (qwen-turn-based + center-out Ascending) & 2155 & `` & 12.09 & 262.42 \\
 \midrule
%-------------------- Appendix (33 rows) -------------------
\multirow{33}{*}{\textbf{Appendix~\ref{sec:rq_appendix}}}
 & Negative pairings (K=1)   & 500 & 13.61  & 10.29 & 281.76 \\
 & Opposite pairings (K=1)   & 500 & 13.34  & 10.04 & 280.69 \\
 & Positive pairings (K=1)   & 500 & 12.87  & 9.72  & 284.83 \\
 & Negative pairings (K=2)   & 992 & 29.34  & 11.67 & 324.25 \\
 & Mixed pairings (K=2)      & 994 & 27.44  & 11.89 & 333.66 \\
 & Positive pairings (K=2)   & 987 & 27.29  & 11.75 & 334.45 \\
 & Negative pairings (K=5)   & 2898 & 70.26 & 14.53 & 390.15 \\
 & Mixed pairings (K=5)      & 2879 & 69.34 & 15.36 & 417.49 \\
 & Positive pairings (K=5)   & 2864 & 67.29 & 14.86 & 423.67 \\
 & Negative pairings (K=10)  & 913  & 151.94& 16.68 & 387.12 \\
 & Mixed pairings (K=10)     & 905  & 150.24& 18.02 & 574.97 \\
 & Positive pairings (K=10)  & 890  & 146.09& 16.56 & 475.81 \\
 & Positive Ratio 0\%        & 125  & ``    & ``    & `` \\
 & Positive Ratio 20\%       & 479  & ``    & ``    & `` \\
 & Positive Ratio 40\%       & 885  & ``    & ``    & `` \\
 & Positive Ratio 60\%       & 869  & ``    & ``    & `` \\
 & Positive Ratio 80\%       & 415  & ``    & ``    & `` \\
 & Positive Ratio 100\%      & 106  & ``    & ``    & `` \\
 & Negative pairings (0.5B)  & 974  & 70.19 & 10.33 & 231.17 \\
 & Mixed pairings (0.5B)     & 970  & 69.32 & 11.55 & 246.13 \\
 & Positive pairings (0.5B)  & 960  & 68.24 & 14.25& 300.61 \\
 & Negative pairings (3B)    & 982  & ``    & 10.81 & 334.58 \\
 & Mixed pairings (3B)       & 984  & ``    & 11.24 & 328.64 \\
 & Positive pairings (3B)    & 985  & ``    & 11.54 & 327.08 \\
 & Negative pairings (7B)    & 2987 & ``    & 10.05 & 393.20 \\
 & Mixed pairings (7B)       & 2916 & ``    & 10.43 & 390.12 \\
 & Positive pairings (7B)    & 2851 & ``    & 10.46 & 374.32 \\
 & Negative pairings (14B)   & 995  & ``    & 12.51 & 333.07 \\
 & Mixed pairings (14B)      & 987  & ``    & 12.68 & 335.39 \\
 & Positive pairings (14B)   & 987  & ``    & 12.21 & 317.31 \\
 & Negative pairings (32B)   & 999  & ``    & 10.83 & 339.31 \\
 & Mixed pairings (32B)      & 980  & ``    & 10.64 & 328.76 \\
 & Positive pairings (32B)   & 959  & ``    & 10.15 & 310.81 \\
\bottomrule
\end{tabular}
} % end resizebox
\caption{Personas and Dialogues Sttistics table combining ConvAI2, synthesized user profiles, and RQ1--RQ4 data.}
\label{tab:statistics}
\end{table*}

Table~\ref{tab:statistics} presents statistics for the ConvAI2 source dataset, the user profiles constructed in \S\ref{sec:userprofile}, and the datasets used in experiments addressing each research question. The symbol "``" indicates that the omitted value deviates only marginally from the reported statistic, while “-” is used when a statistic is either difficult to compute or not applicable. For \S\ref{sec:rq1.1}, only statistics for Llama-3.1-8B are provided for convenience, as its trends are consistent with those observed for \S\ref{sec:rq1.2}.

\end{document}